%% file: arxiv.tex
\documentclass[10pt,twocolumn,letterpaper]{article}

\usepackage{wacv}              

\usepackage{soul}
\usepackage{array}
\usepackage{epsfig}
\usepackage{amsmath}
\usepackage{amssymb}
\usepackage{multicol}
\usepackage{multirow}
\usepackage{booktabs}
\usepackage{graphicx}
\usepackage{xstring}
\usepackage{adjustbox}
\usepackage[dvipsnames]{xcolor}
\usepackage[inline]{enumitem}

\usepackage[pagebackref,breaklinks,colorlinks]{hyperref}

\usepackage[capitalize]{cleveref}
\crefname{section}{Sec.}{Secs.}
\Crefname{section}{Section}{Sections}
\Crefname{table}{Table}{Tables}
\crefname{table}{Tab.}{Tabs.}

\def\meth{UnityGraph\xspace}
\def\embdmeth{Node2Vec\xspace}
\def\cluster{KMeans\xspace}
\def\framework{GPL\xspace}
\def\NumVideos{n\xspace}
\def\SamplingRate{\sigma\xspace}
\def\Stride{\omega\xspace}
\def\WindowSizeUnity{\psi\xspace}
\def\WalkCount{\alpha\xspace}
\def\WindowSizeNodeToVec{\beta\xspace}
\def\WalkLength{\gamma\xspace}
\def\NumFrames{m\xspace}
\def\NumClips{z\xspace}
\def\InputParameter{q\xspace}
\def\ReturnParameter{p\xspace}
\def\EmbedDim{d\xspace}
\newcommand{\myfirstpara}[1]{\noindent \textbf{#1:}}
\newcommand{\mypara}[1]{\vspace{0.2em} \myfirstpara{#1}}
\newcommand{\dt}[1]{\relax}
\newcommand{\mt}[1]{\textcolor{black}{#1}}
\definecolor{self_orange}{HTML}{F46D43}
\definecolor{self_red}{HTML}{D53E4F}
\definecolor{self_purple}{HTML}{5E4FA2}
\definecolor{self_blue}{HTML}{D1E5F0}
\definecolor{self_green}{HTML}{66C2A5}
\definecolor{self_dark_blue}{HTML}{3288BD}

\def\wacvPaperID{236} 
\def\confName{WACV}
\def\confYear{2024}

\begin{document}

\title{United We Stand, Divided We Fall:\\\meth for Unsupervised Procedure Learning from Videos}

\author{Siddhant Bansal\thanks{Corresponding author: \href{mailto:siddhant.bansal@research.iiit.ac.in}{\small siddhant.bansal@research.iiit.ac.in}}\\
CVIT, IIIT, Hyderabad
\and
Chetan Arora\\
IIT, Delhi
\and
C.V. Jawahar\\
CVIT, IIIT, Hyderabad
}
\maketitle

\input{./sections/abstract.tex}

\section{Introduction}\label{introduction}
\input{./sections/introduction.tex}

\section{Related Works}\label{related_works}
\input{./sections/related_works.tex}

\section{Graph-based Procedure Learning (\framework)} \label{methodology}
\input{./sections/meth.tex}

\section{Dataset and Evaluation Methodology}\label{sec:data-n-eval}
\input{./sections/data-n-eval.tex}

\section{Experiments and results}\label{experiments}
\input{./sections/experiments.tex}

\section{Ablation Study}
\input{./sections/ablation.tex}

\section{Conclusion}\label{conclusion}
\input{./sections/conclusion.tex}

\section{Appendix}
This document contains additional results and analysis to support the main paper.
In \Cref{sec:additional_results}, we first compare the results obtained using the proposed \framework framework with CnC for fewer videos.
Then, in \Cref{tab:supp_results_background}, we show the effectiveness of using background frame removal using hand-object interaction.
Finally, in \Cref{sec:tsne_vis}, we visualise and analyse \meth's embeddings obtained using t-SNE~\cite{tsne}.

\section{Additional Results}\label{sec:additional_results}

\mypara{Effect of number of videos to train a method}
In \Cref{tab:supp_increasing_videos}, we compare the representation learned by \framework from CnC~\cite{sid_procel} by increasing the number of videos for the same task.
The table contains results on the Bacon and Eggs category from EgoProceL.
As can be seen in the table, the results of both \framework and CnC increase upon increasing the number of videos.
However, due to the novel \meth -- that creates both spatial and temporal connections -- the results for the less number of videos using the proposed \framework framework are high.
This shows the effectiveness of modeling multiple videos of the same task as a single graph.
On the contrary, CnC~\cite{sid_procel} learns from a pair of videos at a time, leading to learning limited information and hence, low results.

\input{./sections/supp_results_video_count.tex}

\begin{table}[!htpb]
    \centering
    \setlength{\tabcolsep}{5pt}
\caption{\mt{\textbf{Background frame removal percentage}}. \mt{Here we show the amount of frames removed from various EgoProceL's tasks after using Shan} \etal's~\cite{Shan20} \mt{hand-object detector}}
    \begin{tabular}{@{}lr@{}}\toprule
    Category & \% background frames\\
    \midrule
    CMU-MMAC & $18$ \\
    EGTEA G. & $29$ \\
    MECCANO & $4$ \\
    EPIC-Tents & $12$ \\
    PC Assembly & $2$ \\
    PC Disassembly & $2$ \\
    \bottomrule
    \end{tabular}
    \label{tab:background_frame_percent}
\end{table}

\mypara{Effect of Background frame removal}
\input{./sections/supp_results_background.tex}
In \Cref{tab:supp_results_background}, the results obtained with and without background frames on EgoProceL are presented.
As was concluded in the main paper, the results improve for the categories that involve subjects moving around in an unrestricted space.
The reason for this is people working in an unrestricted space need to walk around looking for objects leading to background frames.
As there is no hand-object interaction present in the background frames, using Shan \etal's~\cite{Shan20} hand-object detector, we can filter them.

We observe significant improvement in the results for EGTEA Gaze Plus and EPIC-Tents, as the tasks being performed there are in an unconstrained environment, kitchen and lawn, respectively.
This allows the subjects to search for the objects and wander around during the process.
Also, we observe marginal improvement on CMU-MMAC where the subjects are allowed to work in an unconstrained environment (kitchen) but are constrained by the wires used to capture various modalities.
Due to this, there are on average few sections where the person is not interacting with an object.
On the contrary, no improvements are observed for PC assembly, PC disassembly, and MECCANO.
This is because, here, the subjects are working in a constrained environment -- table top and CPU -- restricting the movement.
Hence, very low number of frames where there is no hand-object interaction.

\mypara{\mt{Amount of background frames removed}}
\Cref{tab:background_frame_percent} \mt{contains the amount of frames removed (in percent) upon using Shan} \etal's \mt{hand-object detector}~\cite{Shan20}.
\mt{It can be seen that the amount of frames without hand-object interaction is proportional to the gain in results (in} \Cref{tab:supp_results_background}).

\section{\meth Embedding Space's Visualisation}\label{sec:tsne_vis}

\input{./sections/supp_fig_tsne_vis.tex}

In this section, we discuss the t-SNE~\cite{tsne} visualisation in \Cref{fig:tsne_vis} generated from \meth's embeddings before and after applying the \embdmeth~\cite{node2vec-kdd2016} algorithm.
On the left-hand side of \Cref{fig:tsne_vis}, t-SNE visualisation for \meth's clip-level embeddings before applying the \embdmeth algorithm are shown.
Though \meth's embeddings are able to capture the background clips well, they fall short of bringing the key-steps close in the embedding space.
As can be seen, the "apply jam" key-step is spread across the embedding space.

On the other hand, on the right-hand side of \Cref{fig:tsne_vis}, t-SNE visualisation for \meth's clip-level embeddings after applying the \embdmeth algorithm is shown.
Here, we can see that the embeddings have arranged themselves in a particular pattern.
In the majority of the cases, the embeddings for similar key-steps have come closer.
For example, the cluster on the top consists of subjects applying peanut butter.
The clips for ``apply jam" have concentrated in a few selected clusters.
And due to the nature of \meth (connecting clips with similar semantic information), background clips with similar styles of information have come closer.
For example, the cluster towards the center has background clips of subjects moving from one location to another.

{\small
\bibliographystyle{ieee_fullname}
\bibliography{bibliography}
}

\end{document}

%% file: sections/abstract.tex
\begin{abstract}
    Given multiple videos of the same task, procedure learning addresses identifying the key-steps and determining their order to perform the task.
    For this purpose, existing approaches use the signal generated from a pair of videos.
    This makes key-steps discovery challenging as the algorithms lack inter-videos perspective.
    Instead, we propose an unsupervised Graph-based Procedure Learning (\framework) framework.
    \framework consists of the novel \meth that represents all the videos of a task as a graph to obtain both intra-video and inter-videos context.
    Further, to obtain similar embeddings for the same key-steps, the embeddings of \meth are updated in an unsupervised manner using the \embdmeth algorithm.
    Finally, to identify the key-steps, we cluster the embeddings using \cluster.
    We test \framework on benchmark ProceL, CrossTask, and EgoProceL datasets and achieve an average improvement of $2\%$ on third-person datasets and $3.6\%$ on EgoProceL over the state-of-the-art.
\end{abstract}

%% file: sections/introduction.tex

\input{./sections/image_intro}

\myfirstpara{Motivation}
Consider developing a robot capable of assembling a phone in a factory.
Hard coding the sequence of steps required to piece together the phone will require years of effort.
Instead, it would be useful if a robot could observe a person fabricating the phone multiple times and learns from it!
Driven by this objective, we focus on unsupervised procedure learning from videos. 
Broadly, the task involves identifying the key-steps and their order via multiple videos of the same activity.

\mypara{Applications}
A framework that can distill the steps required to perform a task from multiple demonstrations could help develop robots for manufacturing pipelines, create assistive machines
,
or monitor and guide a novice learning a new task.
Formally, given multiple videos of a task, procedure learning deals with \textbf{(a)} identifying the key-steps and \textbf{(b)} their order to perform the task~\cite{Shen_action_segmentation_2021_CVPR,multi-task-procl,sid_procel,joint_dynamic_summary}.

\myfirstpara{Difference from action segmentation/detection}
As shown in \Cref{fig:teaser}, procedure learning deals with multiple videos of a task.
In contrast, action-based tasks~\cite{Kumar_2022_CVPR} deal with a single video, hence losing the capability to determine repetitive key-steps across the videos.
Secondly, these tasks do not consider the order of individual events, which is often crucial for identifying key-steps, and/or procedures/recipes. 
For example, action-based tasks do not capture the difference in the order of key-steps in \textcolor{self_dark_blue}{V2} and \textcolor{self_green}{V3} (\cref{fig:teaser}).
Other efforts for video understanding using instructional videos aim at procedure planning~\cite{Zhao_2022_CVPR}, procedure sequence verification~\cite{Qian_2022_CVPR}, and instructional video summarisation~\cite{10.1007/978-3-031-19830-4_31}.
Furthermore, as procedure learning deals with localising the key-steps, it differs from the video alignment task~\cite{tcc,idm}.
Therefore, considering the utility of procedure learning and its distinctness from existing tasks, we devise the Graph-based Procedure Learning (\framework) framework.

\mypara{Our approach}
\framework is a three-staged unsupervised framework for procedure learning.
The first stage of \framework consists of \meth, shown in \Cref{fig:teaser}. 
\meth is a graph that models an arbitrary number of videos from the same task.
For creating \meth at clip-level, the video clips are first passed through a pre-trained I3D ResNet-50~\cite{resnet,Carreira2017QuoVA} to get a \textcolor{self_purple}{node}.
The nodes are later connected based on \textbf{(a)} semantic similarity across videos (\textcolor{self_red}{spatial edges}) and \textbf{(b)} temporal closeness in the same video (\textcolor{self_orange}{temporal edges}).
Due to this structure, \meth captures both the inter-video and intra-videos context.
This sets \meth apart from the previous approaches that estimate the procedure using one~\cite{kukleva2019unsupervised,VidalMata_2021_WACV,multi-task-procl,joint_dynamic_summary,9706996} or two~\cite{sid_procel} videos.

\mypara{Justification of the approach}
As a graph enables us to create a \textcolor{self_red}{spatial edge} between two nodes from different videos irrespective of the key-step order (\Cref{fig:teaser}), it enables us to overcome previous works' key-step ordering constraints~\cite{kukleva2019unsupervised,VidalMata_2021_WACV}.
Also, the range of granularity (number of frames) to create the nodes is controllable, allowing us to test various configurations.
Finally, as shown in \Cref{fig:teaser}, the edges capture two forms of relationships \textbf{(a)} \textcolor{self_orange}{temporal} across the same video and \textbf{(b)} \textcolor{self_red}{semantic} across the videos, enabling us to model inter-video and intra-videos context.

\mypara{Difference from existing techniques}
Most works explore procedure learning in a supervised~\cite{procedure_completion_BMVC_2020,Sener_2019_ICCV,YouCook2} or weakly supervised~\cite{10.1007/978-3-319-10602-1_41,Chang_2019_CVPR,Ding2018WeaklySupervisedAS,Huang2016ConnectionistTM,Li_2019_ICCV,Li_2020_CVPR,8578725,Richard_2018_CVPR,CrossTask} setting.
Supervised methods require frame-level key-step annotations, making them unscalable~\cite{sid_procel}.
On the other hand, weakly supervised learning methods require an ordered or unordered list of key-steps.
Creating the lists requires viewing the videos or defining heuristics leading to scalability issues~\cite{joint_dynamic_summary,multi-task-procl}.
Instead, the second stage of \framework enhances \meth's node embeddings in an unsupervised manner using \embdmeth.

\mypara{Similar works}
The works closely related to ours employ various methods to create frame-level features to identify the procedure.
Kukleva \etal~\cite{kukleva2019unsupervised} use the signal provided by the relative timestamp of the frame. 
Elhamifar \etal~\cite{multi-task-procl} discover and utilise the attention features from individual frames.
Bansal \etal~\cite{sid_procel} solve the problem in a self-supervised manner by utilising the signal from corresponding frames among the videos of the same task.
These works exploit different attributes of videos to extract the signal. 
However, they fall short in creating a graph-based representation to utilise the relationship between all the frames across the input videos. 
In this work, we propose \meth, which first creates a clip-level representation and then captures the correspondences between the key-steps across videos.
%
%
Our \textbf{major contributions} are:
\input{./sections/contributions.tex}

%% file: sections/image_intro.tex
\begin{figure}[!t]
    \centering
    \includegraphics[width=0.4782\textwidth]{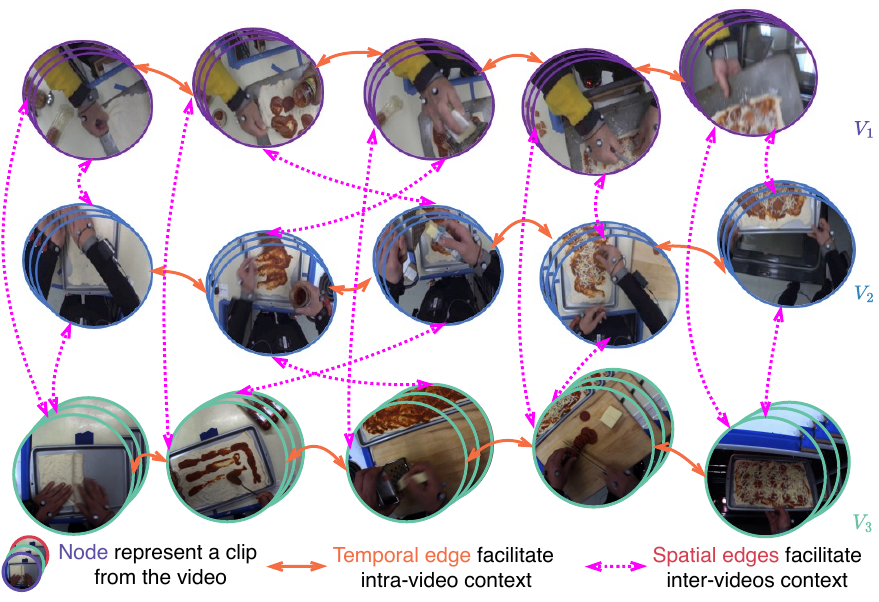}
    \caption{\textbf{\meth} for three pizza making videos. \meth facilitates procedure learning by creating a unified representation of an arbitrary number of videos from the same category.
    Here, the \textcolor{self_purple}{nodes} represent a clip from the video.
    Further, the \textcolor{self_orange}{temporal edges} connect temporally close frames, allowing intra-video context, whereas the \textcolor{self_red}{spatial edges} connect semantically similar frames across the videos, enabling inter-videos context.\vspace{-1.6em}}
    \label{fig:teaser}
\end{figure}

%% file: sections/contributions.tex
\begin{enumerate}[label=(\arabic*),leftmargin=*,topsep=0pt,itemsep=-1ex,partopsep=1ex,parsep=1ex]
    \item We propose the Graph-based Procedure Learning (\framework) framework. Contrary to existing graph-based frameworks, \framework does not require node or edge annotations, enabling unsupervised procedure learning.
    \item We create a novel graph representation for an arbitrary number of videos: \meth. \meth captures \textbf{(a)} temporal relationship in the same video and \textbf{(b)} semantic relationship across the videos.
    \item To identify the background frames, we propose to detect hand-object interactions in egocentric videos. This leads to an improvement of $1.1\%$ on EgoProceL.
    \item We perform experiments on benchmark EgoProceL, ProceL, and CrossTask datasets and achieve on average $2\%$ improvement on third-person datasets (ProceL and CrossTask) and $3.6\%$ improvement on EgoProceL over the state-of-the-art. We will release the code for the work upon acceptance.
\end{enumerate}


%% file: sections/related_works.tex
\myfirstpara{Graph-based Representation for Video Understanding}
Previous works have used graphs for action localisation~\cite{PGCN2019ICCV,Hussein2019VideoGraphRM,xu2020gtad}, task completion~\cite{Huang2019NeuralTG}, video super-resolution~\cite{You_2022_WACV}, grounding in instructional videos~\cite{huang-buch-2018-finding-it}, question answering~\cite{Wang_2021_WACV}, and action recognition~\cite{8953941,chen2019graph,10.1007/978-3-030-01228-1_25,Khan2021SpatiotemporalDM}.
Hussein \etal~\cite{Hussein2019VideoGraphRM} utilise graphs to analyse human activity from a single video.
G-TAD~\cite{xu2020gtad} proposes a graph where video clips are nodes, correlations between the nodes form edges, and actions associated with context are used to create target sub-graphs.
Khan and Cuzzolin~\cite{Khan2021SpatiotemporalDM} propose to create a graph consisting of nodes based on the action tubes and edges encoding the relationships between the action tubes. However, generating action tubes is computationally expensive and not reliable~\cite{Khan2021SpatiotemporalDM}.
Contrary to previous works, \framework proposes \meth to discover key-steps across multiple videos and is the first to utilise graphs for procedure learning.
Furthermore, \meth is a task-level graph compared to the video-level graph in~\cite{xu2020gtad}, enabling us to discover key-steps across videos.
Finally, in contrast to the supervised setting in previous works, \framework learns embeddings in an unsupervised manner.

\input{./sections/meth_figure.tex}

\mypara{Representation Learning for Procedure Learning}
Previous works on procedure learning have developed methods to learn frame-level features~\cite{kukleva2019unsupervised,VidalMata_2021_WACV,joint_dynamic_summary,multi-task-procl}.
Kukleva \etal~\cite{kukleva2019unsupervised} learn the representation space by using relative timestamps of the frames.
On the other hand, Vidal \etal~\cite{VidalMata_2021_WACV} predict the future frame and its timestamps.
Elhamifar \etal~\cite{multi-task-procl} learn and employ attention features for individual frames.
Bansal \etal~\cite{sid_procel} exploit temporal correspondences across the videos to generate the signal and learn frame-level embeddings.
However, these methods fall short in modelling either temporal or spatial relationships.
In contrast, \framework consists of \meth that \textbf{(a)} represents videos at the clip-level and \textbf{(b)} forms edges between semantically similar and temporally close frames.

\myfirstpara{Multi-modal Procedure Learning}
A sub-set of previous works utilises multi-modal data for procedure learning, for example, narrated text and videos~\cite{Inria_dataset,BMVC.28.30,Doughty_2020_CVPR,Fried2020LearningTS,Malmaud2015WhatsCI,Sener_2015_ICCV,Shen_action_segmentation_2021_CVPR,10.1145/2647868.2654997,zhukov20}.
These works have to assume an alignment between the videos and the text~\cite{Inria_dataset,Malmaud2015WhatsCI,10.1145/2647868.2654997}, which is inaccurate in most cases~\cite{multi-task-procl,joint_dynamic_summary}.
Furthermore, they utilise an imperfect Automatic Speech Recognition system to generate text, leading to the requirement of manually cleaning the text, and therefore, is unscalable.
To overcome these issues, following~\cite{sid_procel}, we employ only visual modality as our input to \framework, making our framework highly scalable.

\mypara{Learning Key-step Ordering}
Most of the previous works do not capture different key-step ordering to perform the task.
They either assume strict ordering~\cite{joint_dynamic_summary,kukleva2019unsupervised,VidalMata_2021_WACV} or do not predict the order~\cite{multi-task-procl,Shen_action_segmentation_2021_CVPR}.
However, we observe that subjects perform the same task in multiple ways (\cref{fig:teaser}), motivating us to capture different ways to accomplish a task.
Therefore, \framework aims to create a key-step order for each video and infer the relevant ordering to perform the task.

\mypara{Self-Supervised Representation Learning for Videos}
Recent works have explored various pretext tasks for learning the representation in a self- or unsupervised manner.
For example, utilizing temporal coherence and order as signals~\cite{Fernando2017SelfSupervisedVR,Lee2017UnsupervisedRL,Misra2016ShuffleAL,Choi2020ShuffleAA,Xu_2019_CVPR}, predicting successive frames~\cite{Ahsan2018DiscrimNetSA,Diba2019DynamoNetDA,Han19dpc,Kim2019SelfSupervisedVR,10.5555/3045118.3045209,10.5555/3157096.3157165} or identifying the arrow of time~\cite{8578938}.
Video representation learning methods mentioned generate signals from a limited number of videos. However, our goal is to identify the key-steps from multiple videos.
To this end, we explore graphs for procedure learning and propose \framework consisting of \meth to improve video understanding.

%% file: sections/meth_figure.tex
\begin{figure*}[!thpb]
    \centering
    \includegraphics[width=\textwidth]{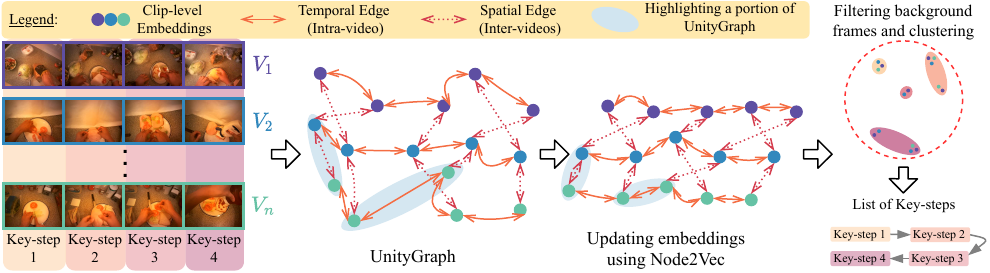}
    \caption{
    \textbf{Graph-based Procedure Learning (\framework) framework.}
    Given multiple videos of the same task, we create \meth.
    Using the \embdmeth algorithm, we exploit the structure of \meth to enhance the node embeddings in an unsupervised manner.
    For example, the temporal and spatial clips that were originally far in the embedding space are closer after \embdmeth (highlighted in \textcolor{self_blue}{blue}).
    Finally, we cluster the embeddings using \cluster and filter the background frames to obtain the key-steps required to perform the task.
    }
    \label{fig:meth_procegraph}
\end{figure*}

%% file: sections/meth.tex

Autonomously inferring the key-steps required to perform a task opens up the possibility of creating a variety of autonomous, guidance, and assistive systems.
The majority of the previous works generate self-supervised signals from either single or a couple of videos.
However, to better discover the procedure, getting the signal across all the videos is crucial.
To this end, we utilise the capability of graphs to represent abstract video data.
As shown in \Cref{fig:meth_procegraph}, the first part of \framework framework consists of the proposed \meth.
It is a novel graph representation for an arbitrary number of videos of a task (\Cref{sec:unity_graph_creation}).

The initial features of \meth are created using a pre-trained I3D ResNet-50~\cite{resnet,Carreira2017QuoVA}.
To further improve the features, as shown in \Cref{fig:meth_procegraph}, the next step in \framework involves updating the embeddings using the \embdmeth algorithm in an unsupervised manner.
Once the embeddings are learned, they are clustered using the \cluster algorithm (\Cref{sec:learning_clustering_embeddings}).
The final step of \framework involves ordering the discovered clusters based on the average timestamps of the constituting frames (\Cref{sec:learning_clustering_embeddings}).

\mypara{Notations}
As shown in \Cref{fig:meth_procegraph}, \framework takes in $\NumVideos$ untrimmed videos of the same task, denoted by $V=\{V_i: i \in \mathbb{N}, 1 \leq i \leq \NumVideos\}$.
Note that the $\NumVideos$ videos can have a different number of frames.
A video $V_x$ with $\NumFrames$ frames is divided into multiple clips using a sampling rate, stride, and window size of $\SamplingRate$, $\Stride$, and $\WindowSizeUnity$, respectively.
The clips are then passed through a pre-trained I3D ResNet-50, denoted as $f_\theta$ (with parameters $\theta$), used to generate node-level embeddings of dimension $\EmbedDim$ for \meth.
The clips for video $V_x$ with $\NumClips$ clips are denoted as $V_x = \{c_x^1, c_x^2, \dots, c_x^\NumClips\}$ and the video's node-level embeddings are denoted as $f_{\theta}(V_x) = \{v_x^1, v_x^2, \dots, v_x^\NumClips\}$.
Furthermore, we assume $K$ key-steps in a task where $K$ is a hyperparameter.

\subsection{Representing Videos using \meth}\label{sec:unity_graph_creation}

\mypara{Assumptions}
We make the following design choices when creating \meth:
\begin{enumerate*}[label={\textbf{(\alph*)}}]
    \item To compensate for the high frame rate and long action duration, we create \meth's nodes at the clip level.
    \item Using a 3D CNN, each clip is converted to an embedding.
    The motivation here is that the sampled clip either contains one action or none.
    \item To keep the problem tractable, we assume the objects and actions are semantically similar across the task's videos.
\end{enumerate*}

\subsubsection{Creating \meth's nodes and edges}\label{sec:node_edges_creation}

\input{./sections/meth_figure_graph_creation.tex}

\Cref{fig:unitygraph_creation} summarises the creation of \meth.
A node $v_x^1$ in \meth is the embedding of a clip $c_x^1$ for video $V_x$.
The embedding is a $\EmbedDim$ dimensional vector created using an I3D ResNet-50~\cite{resnet,Carreira2017QuoVA}.
For example, for \textcolor{self_purple}{$V_1$} with $\NumClips = 5$, the nodes are created as:

\begin{equation}\label{equ:node_equation}
\textcolor{self_purple}{v_1^i} = f_\theta(\textcolor{self_purple}{c_1^i}),  \text{where} \; i \in \{1, \dots, \NumClips\}
\end{equation}

Creating nodes in this way helps with
\begin{enumerate*}[label={\textbf{(\alph*)}}]
    \item converting a volume of frames (the clip) to an embedding and
    \item comparing and modifying the embeddings to understand the procedure.
\end{enumerate*}
\Cref{fig:unitygraph_creation} (a) summarises this process.

Once the nodes are created, the graph is completed by creating edges between them.
The edges are created at two levels \begin{enumerate*}[label={\textbf{(\alph*)}}]
\item \textcolor{self_red}{spatial} that facilitate inter-videos connection and 
\item \textcolor{self_orange}{temporal} that facilitate intra-video connection.
\end{enumerate*}

To better understand the process, consider two videos (\textcolor{self_purple}{$V_1$}, \textcolor{self_dark_blue}{$V_2$})  from \Cref{fig:unitygraph_creation} (b).
Let us focus on creating an edge between the first node (\textcolor{self_dark_blue}{$v_2^1$}) from \textcolor{self_dark_blue}{$V_2$} and nodes from \textcolor{self_purple}{$V_1$}.
The goal is to find the node in \textcolor{self_purple}{$V_1$} having the highest semantic similarity with \textcolor{self_dark_blue}{$v_2^1$}.
To this end, we calculate the cosine similarity ($S_C$) between \textcolor{self_dark_blue}{$v_2^1$} and all the nodes in \textcolor{self_purple}{$V_1$}:
\begin{align}\label{equ:cosine_similarity}
    S_C(\textcolor{self_dark_blue}{v_2^1}, \textcolor{self_purple}{v_1^i}) = \frac{\sum_{j=1}^\EmbedDim \textcolor{self_dark_blue}{v_{2j}^1} \textcolor{self_purple}{v_{1j}^i}}{\sqrt{\sum_{j=1}^\EmbedDim (\textcolor{self_dark_blue}{v_{2j}^{1}})^2}\sqrt{\sum_{j=1}^\EmbedDim (\textcolor{self_purple}{v_{1j}^{i}})^2}}, 
    \nonumber \\ \text{where} \; i \in \{1, \dots, z\}.
\end{align}
The \textcolor{self_red}{spatial edge} is created between the node with the highest similarity score:
\begin{align}\label{equ:spatial_edge}
    \text{\textcolor{self_red}{Edge}}(\textcolor{self_dark_blue}{v_2^1}, \textcolor{self_purple}{v_1^i}) = 
    \begin{cases}
    1, \text{if} \max(S_C(\textcolor{self_dark_blue}{v_2^1}, \textcolor{self_purple}{v_1^i})) 
    \\ 0, \text{otherwise}
    \end{cases},
    \nonumber\\ \text{where} \; i \in \{1, \dots, z\}.
\end{align}
To create the \textcolor{self_orange}{temporal edges}, we connect the neighboring nodes from the same video.
Let us consider creating \textcolor{self_orange}{temporal edges} for \textcolor{self_dark_blue}{$V_2$}: 
\begin{align}\label{equ:temporal_edge}
    \text{\textcolor{self_orange}{Edge}}(\textcolor{self_dark_blue}{v_2^i}, \textcolor{self_dark_blue}{v_2^j}) = 
    \begin{cases}
    1, \text{if} \; |i - j| = 1
    \\ 0, \text{otherwise}
    \end{cases},
    \nonumber \\ \text{where} \; i, j \in \{1, \dots, z\}.
\end{align}
To summarise, \meth consists of nodes created using \Cref{equ:node_equation}. The nodes are \textcolor{self_red}{spatially} connected using \Cref{equ:spatial_edge} and \textcolor{self_orange}{temporally} connected using \Cref{equ:temporal_edge}.

\subsubsection{Detecting background frames}\label{sec:background_frames}

\begin{figure}[t]
    \centering
    \includegraphics[width=0.4782\textwidth]{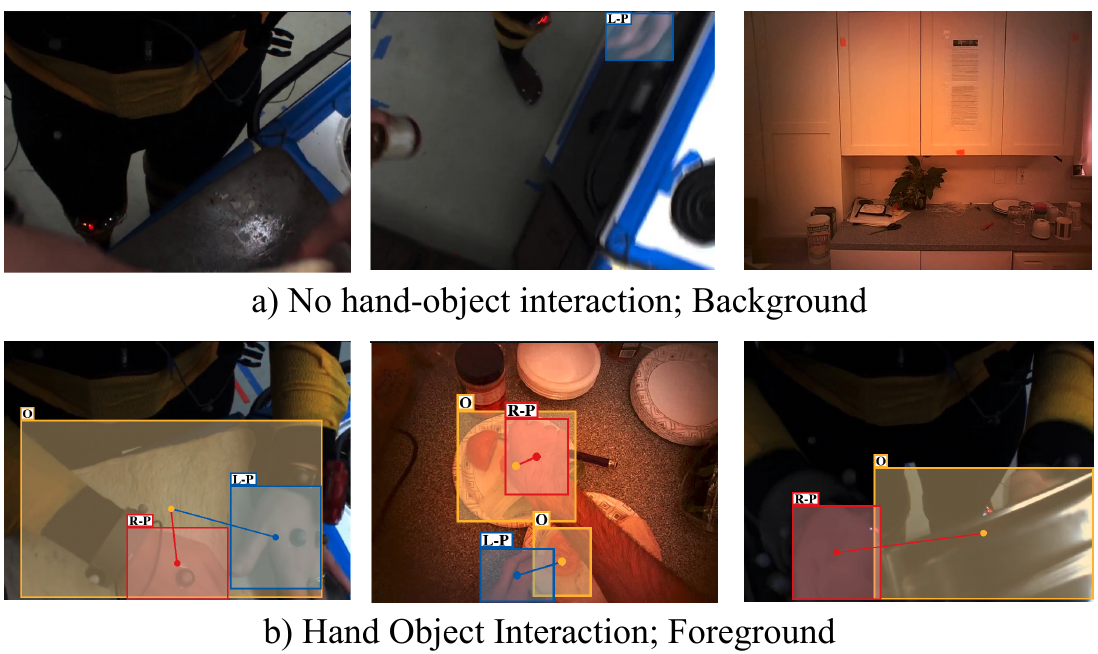}
    \caption{We use the hand-object detection model from~\cite{Shan20}. a) Frames not containing hand-object interaction. Second image in the first row contains a hand without an interaction with an object, hence, background. b) Frames containing hand-object interaction and contribute towards understanding the procedure.}
    \label{fig:hand_obj_interaction}
\end{figure}

The procedure learning datasets majorly consists of background frames~\cite{CrossTask,sid_procel}, making it difficult to determine the procedure.
We observe that a majority of the background frames involve people searching for objects, reading instructions, and waiting for an automated step to finish.
Furthermore, as shown in \cref{fig:hand_obj_interaction}, these activities do not involve hand-object interaction.
Therefore, we argue that the frames lacking hand-object interactions represent the background and we propose to use Shan \etal's hand-object interaction model~\cite{Shan20} to filter out such frames in egocentric videos.

\input{./sections/table_hyperparameters.tex}

\subsection{Identifying Key-steps and their Order}\label{sec:learning_clustering_embeddings}

\myfirstpara{Updating and Clustering \meth's Embeddings}
As illustrated in \Cref{fig:meth_procegraph}, using default embeddings obtained from the pre-trained network can result in embeddings lying far from each other.
To improve the embeddings in an unsupervised manner, we update them using the Node2Vec algorithm~\cite{node2vec-kdd2016}.
The updated embeddings are clustered using \cluster to discover the key-steps.

\mypara{Utility of Node2Vec} 
Node2Vec~\cite{node2vec-kdd2016} learns embeddings while preserving neighborhood information by simulating biased random walks.
UnityGraph connects semantically similar frames from multiple videos and temporally close frames from the same video.
Node2Vec exploits these connections (refer to Figure 1 in supplementary) in an unsupervised manner to improve the embeddings.
In contrast, DeepWalk~\cite{deepwalk} utilises uniform random walks and falls short of capturing this structure~\cite{node2vec-kdd2016}.

\mypara{Identifying the Order of Key-steps}
Once we have the clusters of key-steps, we follow~\cite{sid_procel} to determine their order.
For each clip, the normalized time is calculated~\cite{kukleva2019unsupervised,sid_procel}.
Based on the cluster clips' normalized time, the average time for the cluster is calculated.
The clusters are then arranged in an increasing order of time to generate the order of key-steps.
This approach has two advantages \begin{enumerate*}[label={\textbf{(\alph*)}}] \item it allows each video to have its own key-step order and \item does not require providing key-step ordering information.
\end{enumerate*}

\mypara{Complexity Analysis}
We extract information simultaneously from multiple offline videos and do not optimise for time.
Hence, GPL's time complexity is exponential in the number of videos.
Considering we have $n$ videos of the same length ($l$), then the time complexity is $O(n^2l^2)$.

%% file: sections/meth_figure_graph_creation.tex
\begin{figure}[!tphb]
    \centering
    \includegraphics[width=0.4782\textwidth]{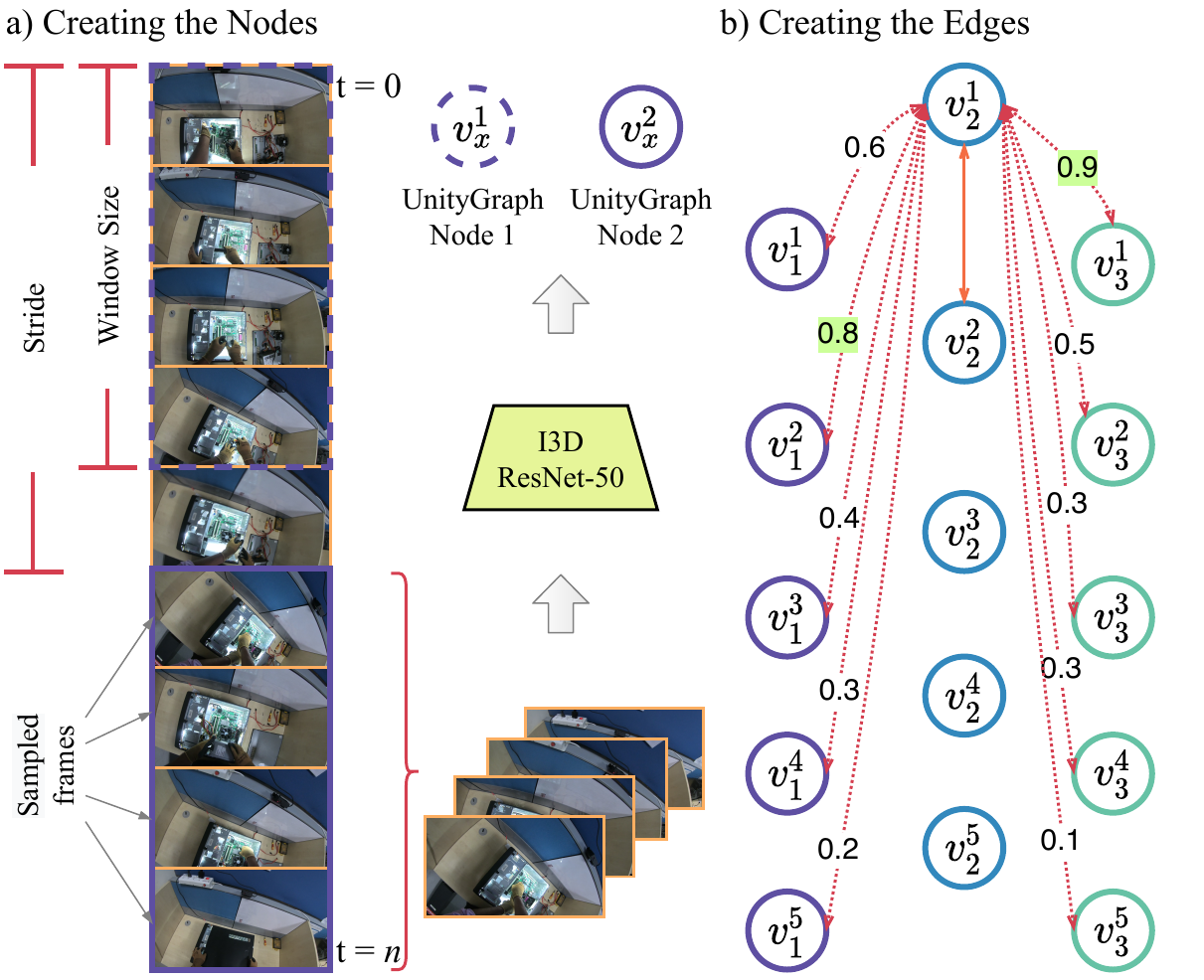}
    \caption{a) Given window size ($\WindowSizeUnity$), stride ($\Stride$), and sampling rate ($\SamplingRate$), a clip from a video is passed through a pre-trained I3D ResNet-50 to generate the node's embedding.
    b) We consider nodes from three videos (\textcolor{self_purple}{$V_1$}, \textcolor{self_dark_blue}{$V_2$}, \textcolor{self_green}{$V_3$}).
    For brevity, we show similarity score between \textcolor{self_dark_blue}{$v_2^1$} and all the nodes in \textcolor{self_purple}{$V_1$} and \textcolor{self_green}{$V_3$}.
    The edges with highest semantic similarity (marked in green) are retained.
    }
    \label{fig:unitygraph_creation}
\end{figure}

%% file: sections/table_hyperparameters.tex
\begin{table}[t]
    \centering
    \caption{\textbf{Hyper-parameter values} for different components of the \framework framework.
    Here, ``\textbf{Ego}'' refers to first-person, ``\textbf{ThP}'' refers to third-person, and 
    ``Ablation table" refers to the table containing quantitative results for the respective hyper-parameter
    \vspace{-1em}
    }
    \begin{adjustbox}{width=0.475\textwidth}
    \begin{tabular}{@{}lrrrr@{}}\toprule
        \multirow{2}{1.5cm}{Hyper-parameter} & Notation & Value (\textbf{Ego}) & Value (\textbf{ThP}) & \multirow{2}{1.2cm}{Ablation Table} \\
        \\
        \midrule
        Sampling Rate & $\SamplingRate$ & $8$ & $4$ & \Cref{tab:unity_graph_hyperparameters_ablation}\\
        Stride & $\Stride$ & $5$ & $10$ &  \Cref{tab:unity_graph_hyperparameters_ablation}\\
        \multirow{2}{1.9cm}{\meth's Window Size} &  \multirow{2}{*}{$\WindowSizeUnity$} &  \multirow{2}{*}{$10$} &  \multirow{2}{*}{$10$} &  \multirow{2}{*}{\Cref{tab:unity_graph_hyperparameters_ablation}}\\
        \\
        Walk Count & $\WalkCount$ & $100$ & $50$ & \Cref{tab:embedding_creation_ablation}\\
        Walk Length & $\WalkLength$ & $100$ & $50$ &  \Cref{tab:embedding_creation_ablation}\\
        \multirow{2}{1.9cm}{\embdmeth's Window Size} &  \multirow{2}{*}{$\WindowSizeNodeToVec$} &  \multirow{2}{*}{$12$} &  \multirow{2}{*}{$10$} &   \multirow{2}{*}{\Cref{tab:embedding_creation_ablation}}\\
        \\
        Return Parameter & $\ReturnParameter$ & $1.0$ & $1.0$ & \Cref{tab:graph_walk_ablation}\\
        In-out Parameter & $\InputParameter$ & $1.0$ & $1.0$ & \Cref{tab:graph_walk_ablation}\\
        \multirow{2}{1.5cm}{No. of Key-steps} & \multirow{2}{*}{$K$} & \multirow{2}{*}{$7$} & \multirow{2}{*}{$7$} & \multirow{2}{*}{\Cref{tab:K_ablation}}\\
        \\
        \multirow{2}{1.5cm}{No. of Videos} & \multirow{2}{*}{$\NumVideos$} & \multirow{2}{*}{$\max(\NumVideos)$} & 
        \multirow{2}{*}{$\max(\NumVideos)$} & \multirow{2}{*}{\Cref{tab:increasing_videos}}\\
        \\
        \multirow{2}{1.5cm}{Embedding Dimension} &  \multirow{2}{*}{$\EmbedDim$} &  \multirow{2}{*}{$400$} &  \multirow{2}{*}{$400$} &  \multirow{2}{*}{$-$}\\
        \\
        \bottomrule
    \end{tabular}
    \end{adjustbox}
    \vspace{-1.3em}
    \label{tab:hyperparameters}
\end{table}


%% file: sections/data-n-eval.tex
\myfirstpara{Evaluation} 
Unless otherwise mentioned, we evaluate the proposed \framework framework following~\cite{sid_procel}.
We take the mean of the scores over all the key-steps and report F1 and IoU Scores.
F1-Score is the harmonic mean of the precision and recall scores.
For precision, we calculate the ratio between the number of frames having correct key-steps prediction and the number of frames assigned to the key-steps.
For recall, the denominator is the number of ground truth key-step frames across all the key-steps of the video.
Following~\cite{Inria_dataset,multi-task-procl,joint_dynamic_summary,Shen_action_segmentation_2021_CVPR,kukleva2019unsupervised,sid_procel}, we obtain the one-to-one mapping between the ground truth and prediction using the Hungarian algorithm~\cite{hungarian_algorithm}.

\input{./sections/results_third_person.tex}
\mypara{Experimental Setup}\label{experimental_setup}
We use features from the final layer of 3D ResNet-50~\cite{resnet} pre-trained on Kinetics 400~\cite{Carreira2017QuoVA} provided by PyTorch~\cite{PyTorch}.
To keep feature extraction tractable, we reshape the short side of the video frame to $256$, while maintaining the aspect ratio.
For detecting the hand-object interactions, we use the `handobj\_100K+ego' model provided by~\cite{Shan20}.
We create and manipulate graphs using NetworkX~\cite{networkx}.
Furthermore, \Cref{tab:hyperparameters} contains the hyper-parameter values obtained for egocentric and third-person view after an extensive ablation study.

\input{./sections/results_egocentric.tex}

\mypara{Baselines}\label{baselines}
\begin{enumerate*}[label={\textbf{(\alph*)}}]
	\item \textbf{Random}: Here, the labels are obtained by randomly sampling predictions from a uniform distribution with $K$ values representing $K$ key-steps.
	\item \textbf{CnC}~\cite{sid_procel}: This work generates frame-level embeddings by learning an embedding space that exploits temporal correspondences across a couple of videos.
    \item \textbf{\framework-2D}: Here, to compare between clip- and frame-level features, we create \meth nodes utilizing features from ResNet-50 initialised on ImageNet and further use Node2Vec to update the features.
	\item \textbf{UG-I3D}: Here, we do not update the embeddings using \embdmeth. Instead, \textbf{U}nity\textbf{G}raph consisting of nodes embeddings from I3D ResNet-50~\cite{resnet,Carreira2017QuoVA}.
\end{enumerate*}

\mypara{Datasets}\label{datasets}
Contrary to previous works that use either first- or third-person datasets for procedure learning, we perform experiments on both views.
For third-person procedure learning, we choose standard benchmark datasets, CrossTask~\cite{CrossTask} and ProceL~\cite{joint_dynamic_summary}.
CrossTask consists of $213$ hours of videos from $18$ primary tasks ($2763$ videos).
ProceL consists of $47.3$ hours of videos from $12$ diverse tasks ($720$ videos).
To demonstrate the efficiency of the proposed \framework framework, we evaluate it in the first-person EgoProceL~\cite{sid_procel} dataset.
It consists of $62$ hours of egocentric videos of $130$ subjects performing $16$ tasks.
\mt{Please note that we \textit{do not} alter the key-step ordering provided by these datasets.
Hence, they are identical to the previous works }~\cite{sid_procel,kukleva2019unsupervised,multi-task-procl,Shen_action_segmentation_2021_CVPR,Inria_dataset,Fried2020LearningTS}.

%% file: sections/results_third_person.tex
\begin{table}[t]
    \centering
    \caption{\textbf{Procedure Learning from Third-person Videos}. Comparison between state-of-the-art methods and \framework on third-person datasets~\cite{joint_dynamic_summary,CrossTask}. 
    Results in \textbf{bold} and \underline{underline} are the highest and second highest in a column, respectively.
    \textbf{P}, \textbf{R}, and \textbf{F} represent precision, recall, and F-score, respectively\vspace{-1em}
    }
    \begin{adjustbox}{width=0.475\textwidth}
    \begin{tabular}{@{}lrrrrrr@{}}\toprule
        & \multicolumn{3}{c}{ProceL~\cite{joint_dynamic_summary}} & \multicolumn{3}{c}{CrossTask~\cite{CrossTask}}\\
        \cmidrule(lr){2-4}\cmidrule(l){5-7}
        & \textbf{P} & \textbf{R} & \textbf{F} & \textbf{P} & \textbf{R} & \textbf{F} \\
        \midrule
        Uniform & $12.4$ & $9.4$ & $10.3$ & $8.7$ & $9.8$ & $9.0$\\
        Alayrc \etal~\cite{Inria_dataset} & $12.3$ & $3.7$ & $5.5$ & $6.8$ & $3.4$ & $4.5$ \\
        Kukleva \etal~\cite{kukleva2019unsupervised} & $11.7$ & $\underline{30.2}$ & $16.4$ & $9.8$ &  $\underline{35.9}$& $15.3$ \\
        Elhamifar \etal~\cite{multi-task-procl} & $9.5$ & $26.7$ & $14.0$ & $10.1$ & $\mathbf{41.6}$ & $16.3$ \\
        Fried \etal~\cite{Fried2020LearningTS} & $-$ & $-$ & $-$ & $-$ & $28.8$ & $-$ \\
        Shen \etal~\cite{Shen_action_segmentation_2021_CVPR} & $16.5$ & $\mathbf{31.8}$ & $21.1$ & $15.2$ & $35.5$ & $21.0$  \\
        CnC~\cite{sid_procel} & $20.7$ & $22.6$ & $21.6$ & $22.8$ & $22.5$ & $22.6$ \\
        \framework-2D (\textit{ours}) & $\underline{21.7}$ & $23.8$ & $\underline{22.7}$ & $\underline{24.1}$ & $23.6$ & $\underline{23.8}$ \\
        UG-I3D (\textit{ours}) & $21.3$ & $23.0$ & $22.1$ & $23.4$ & $23.0$ & $23.2$ \\
        \framework (\textit{ours}) & $\mathbf{22.4}$ & $24.5$ & $\mathbf{23.4}$ & $\mathbf{24.9}$ & $24.1$ & $\mathbf{24.5}$ \\
        \bottomrule\vspace{-2.5em}
    \end{tabular}
    \end{adjustbox}
    \label{tab:table_third_person_results}
\end{table}

%% file: sections/results_egocentric.tex
\begin{table*}[t]
    \centering
    \setlength{\tabcolsep}{7pt}
    \caption{\textbf{Results on egocentric view} on EgoProceL,
    \framework outperforms previous work on most of the tasks.
    This highlights the effectiveness of video representation generated using the proposed \meth and \embdmeth for updating the embeddings based on node neighborhoods.
    Note that EgoProceL is a recent dataset for egocentric procedure learning, due to this, there is only one approach (CnC~\cite{sid_procel}) to fairly compare with.
    Furthermore, as other methods have been specifically designed around third-person datasets, we compare with them on those datasets in 
    \Cref{tab:table_third_person_results}.
    Here, CnC and GPL-2D (with Node2Vec) utilize features from ResNet-50 initialised on ImageNet whereas, UG-I3D (without Node2Vec) and GPL (with Node2Vec) utilize features from I3D ResNet-50 initialised on Kinetics-400.
    Results in \textbf{bold} and \underline{underline} are the highest and second highest in a column, respectively\vspace{-1em}
    }
    \begin{tabular}{@{}lcccccccccccc@{}}
    \toprule
    & \multicolumn{12}{c}{EgoProceL}\\
    \cmidrule{2-13}
    & \multicolumn{2}{c}{CMU-MMAC} & \multicolumn{2}{c}{EGTEA G.} & \multicolumn{2}{c}{MECCANO} & \multicolumn{2}{c}{EPIC-Tents} & \multicolumn{2}{c}{PC Assembly} & \multicolumn{2}{c}{PC Disassembly} \\
    \cmidrule(lr){2-3}\cmidrule(lr){4-5}\cmidrule(lr){6-7}\cmidrule(lr){8-9}\cmidrule(lr){10-11}\cmidrule(l){12-13}
    & F1 & IoU & F1 & IoU & F1 & IoU & F1 & IoU & F1 & IoU & F1 & IoU \\
    \midrule
    Random & $15.7$ & $5.9$ & $15.3$ & $4.6$ & $13.4$ & $5.3$ & $14.1$ & $6.5$ & $15.1$ & $7.2$ & $15.3$ & $7.1$\\
    CnC~\cite{sid_procel} & $22.7$ & $11.1$ & $21.7$ & $9.5$ & $18.1$ & $7.8$ & $17.2$ & $8.3$ & $\underline{25.1}$ & $\underline{12.8}$ & $\underline{27.0}$ & $14.8$\\
    \framework-2D (\textit{ours}) & $21.8$ & $11.7$ & $23.6$ & $14.3$ & $18.0$ & $\underline{8.4}$ & $\underline{17.4}$ & $\underline{8.5}$ & $24.0$ & $12.6$ & $\mathbf{27.4}$ & $\textbf{15.9}$ \\
    UG-I3D (\textit{ours}) & $\underline{28.4}$ & $\underline{15.6}$ & $\underline{25.3}$ & $\underline{14.7}$ & $\underline{18.3}$ & $8.0$ & $16.8$ & $8.2$ & $22.0$ & $11.7$ & $24.2$ & $13.8$ \\ 
    \framework (\textit{ours}) & $\mathbf{31.7}$ & $\mathbf{17.9}$ & $\mathbf{27.1}$ & $\mathbf{16.0}$ & $\mathbf{20.7}$ & $\mathbf{10.0}$ & $\mathbf{19.8}$ & $\mathbf{9.1}$ & $\mathbf{27.5}$ & $\mathbf{15.2}$ & $26.7$ & $\underline{15.2}$ \\
    \bottomrule
    \end{tabular}
    \label{tab:results_first_person}
\end{table*}

%% file: sections/experiments.tex
\myfirstpara{Results on Third-person View}
\Cref{tab:table_third_person_results} compares state-of-the-art methods and \framework on two third-person datasets~\cite{CrossTask,joint_dynamic_summary}.
We obtain the results for the previous works from~\cite{Shen_action_segmentation_2021_CVPR,sid_procel}.
Here, for a fair comparison, the framework is evaluated using the metrics laid out in~\cite{kukleva2019unsupervised,multi-task-procl,Shen_action_segmentation_2021_CVPR}.
\mt{We observe that}~\cite{kukleva2019unsupervised,multi-task-procl} \mt{assigns majority of the frames to one key-step. Due to this, the recall is high, however, the precision decreases, lowering the F-score significantly}.

\mypara{Results on Egocentric View}\label{first_person_results}
\Cref{tab:results_first_person} compares state-of-the-art and \framework on the EgoProceL dataset.
The results for tasks in CMU-MMAC~\cite{CMU_Kitchens} and EGTEA G.~\cite{egtea_gaze_p} have been averaged and reported.
Note that EgoProceL is a recent dataset for egocentric procedure learning, hence, there is only one approach~\cite{sid_procel} to fairly compare with.
We compare methods developed on third-person datasets in \Cref{tab:table_third_person_results}.

The results obtained highlight \begin{enumerate*}[label=\textbf{(\alph*)}]
\item generalisation capabilities of \framework. 
As \framework obtains high results on tasks using objects with high variability in size, different locations, and a variety of lighting conditions.
\item Effectiveness of using \meth for modeling temporal and spatial relationships across the videos by creating a single representation for an arbitrary number of videos.
\item 
The results consistently increase upon using \embdmeth for updating the embeddings of UG-I3D.
This further justifies our hypothesis, shown in \Cref{fig:meth_procegraph}, that \embdmeth improves the embeddings and helps in inferring the procedure.
\item Utility of using clips for creating \meth.
\framework-2D performs comparable to previous frame-level techniques (\eg~\cite{sid_procel}). 
However, as clips allow averaging the frame-level noise, collating information from clips performs the best.
\end{enumerate*}

\mypara{Qualitative Results}\label{qualitative_results}
\input{./sections/qualitative_results}
\Cref{fig:qualitative_results} shows the qualitative results obtained using the baselines and the proposed \framework framework on two tasks from EgoProceL.

%% file: sections/qualitative_results.tex
\begin{figure*}[!tphb]
    \centering
    \includegraphics[width=\textwidth]{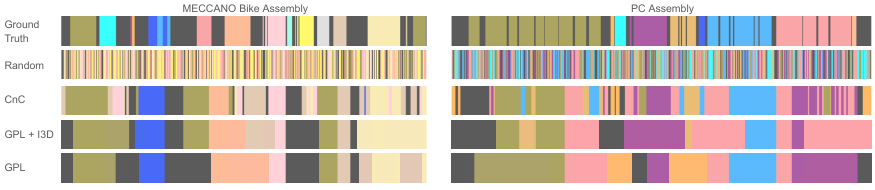}
    \caption{\textbf{Qualitative Results} for one video each of Bike and PC Assembly.
    Each color for a task denotes one key-step and gray sections are the background.
    First row contains the ground truth label, second row contains the results obtained by randomly predicting the key-steps, third row shows results obtained using CnC~\cite{sid_procel}, fourth row highlight the results using \meth's node generated using I3D ResNet-50, and the last row shows results obtained for the \framework framework.
    As can be seen, the segments obtained from \framework are more coherent upon using \embdmeth to update \meth's embeddings.
    This highlights the efficacy of both, \meth and \embdmeth
    }
    \label{fig:qualitative_results}
\end{figure*}

%% file: sections/ablation.tex
To determine optimal hyper-parameters, unless otherwise mentioned, we perform ablation on two challenging tasks: PC Assembly \dt{(egocentric)}~\cite{sid_procel} and Change Tire \dt{(third-person)}~\cite{joint_dynamic_summary}.
\dt{Furthermore,} Due to different attributes of first- and third-person videos, we select one set of hyper-parameters for each of the views.
Finally, unless otherwise mentioned, for egocentric videos, \meth is created using background frame detection.

\mypara{Creating \meth}\label{creating_unity_graph_ablation}
\input{./sections/unitygraph_ablation.tex}

\mypara{Learning and clustering the embeddings}\label{learning_clustering_embeddings_ablation}
\input{./sections/embeddings_ablation.tex}

\mypara{Number of key-steps and background frames}\label{sec:background_frames_ablation}
\input{./sections/procedure_learning_ablation.tex}

\mypara{Number of videos for creating \meth}\label{sec:num_videos_ablation}
\input{./sections/video_count_ablation.tex}

%% file: sections/unitygraph_ablation.tex
\begin{table}[t]
    \centering
    \caption{\textbf{Hyper-parameters for creating \meth}. Here, the results are obtained upon changing various parameters for creating \meth. \textbf{R}, and \textbf{F} represent recall, and F-score, respectively\vspace{-1em}}
    \begin{adjustbox}{width=0.475\textwidth}
    \begin{tabular}{@{}lllcccccc@{}}\toprule
    \multirow{2}{1.2cm}{Sampling Rate} & \multirow{2}{0.8cm}{Stride} & \multirow{2}{1cm}{Window Size} & \multicolumn{3}{c}{PC Assembly} & \multicolumn{3}{c}{Change Tire}\\
        \cmidrule(lr){4-6}\cmidrule(l){7-9}
    & & & \textbf{R} & \textbf{F} & \textbf{IoU} & \textbf{R} & \textbf{F} & \textbf{IoU} \\
        \midrule
$4$ & $5$ & $10$ & $23.0$ & $22.6$ & $12.5$ & $23.1$ & $20.7$ & $12.6$ \\
    $8$ & $5$ & $10$ & $\mathbf{28.8}$ & $\mathbf{27.2}$ & $\mathbf{15.1}$ & $23.8$ & $21.0$ & $12.8$ \\
    $4$ & $10$ & $10$ & $20.0$ & $19.9$ & $11.0$ & $\mathbf{26.2}$ & $\mathbf{23.2}$ & $\mathbf{13.9}$ \\
    $8$ & $10$ & $10$ & $28.1$ & $26.8$ & $15.0$ & $23.7$ & $21.1$ & $12.7$ \\
    $4$ & $15$ & $10$ & $25.0$ & $24.3$ & $13.5$ & $23.8$ & $21.4$ & $13.1$ \\
    $8$ & $15$ & $10$ & $25.7$ & $26.6$ & $14.3$ & $23.2$ & $21.0$ & $12.9$ \\
    $4$ & $5$ & $15$ & $22.2$ & $21.7$ & $11.9$ & $24.1$ & $21.0$ & $12.4$ \\
    $8$ & $5$ & $15$ & $21.1$ & $20.6$ & $10.7$ & $21.2$ & $19.6$ & $11.9$ \\
    $4$ & $10$ & $15$ & $25.2$ & $24.0$ & $13.2$ & $23.9$ & $21.6$ & $13.2$ \\
    $8$ & $10$ & $15$ & $23.4$ & $22.3$ & $12.3$ & $25.1$ & $22.8$ & $13.6$ \\
    $4$ & $15$ & $15$ & $23.7$ & $22.9$ & $12.8$ & $22.7$ & $20.1$ & $12.2$ \\
    $8$ & $15$ & $15$ & $22.7$ & $21.8$ & $11.9$ & $24.2$ & $21.8$ & $13.3$ \\
        \bottomrule\vspace{-1.5em}
    \end{tabular}
    \end{adjustbox}
    \label{tab:unity_graph_hyperparameters_ablation}
\end{table}
In \Cref{tab:unity_graph_hyperparameters_ablation}, we \mt{check} \dt{test various} values for frame sampling rate ($\SamplingRate$), stride ($\Stride$), and window size ($\WindowSizeUnity$).
As egocentric videos have high motion variability, the
maximum scores are obtained for a high sampling rate of $8$.
Also, the stride ($5$) and window size ($10$) are lowest for egocentric videos enabling the creation of nodes with high variability in less time.
For third-person videos, the maximum scores are obtained for a low sampling rate of $4$.
This is because third-person cameras are fixed and do not have high variability in scenes.
Furthermore, the stride of $10$ and window size of $10$ works best as it allows to sample sparsely.
%
%
\dt{Table 5 has results obtained after using various similarity measurement approaches for creating edges of UnityGraph.}

%% file: sections/embeddings_ablation.tex
\begin{table}[t]
    \centering
    \caption{\textbf{Hyper-parameters for learning the embeddings}. Here, the results are obtained upon varying \embdmeth's parameters. \textbf{R}, and \textbf{F} represent recall, and F-score, respectively\vspace{-1em}}
    \begin{adjustbox}{width=0.475\textwidth}
    \begin{tabular}{@{}lllcccccc@{}}\toprule
\multirow{2}{1cm}{Walk Count} & \multirow{2}{1cm}{Walk Length} & \multirow{2}{1.2cm}{Window Size} & \multicolumn{3}{c}{PC Assembly} & \multicolumn{3}{c}{Change Tire}\\
        \cmidrule(lr){4-6}\cmidrule(l){7-9}
& & & \textbf{R} & \textbf{F} & \textbf{IoU} & \textbf{R} & \textbf{F} & \textbf{IoU} \\
        \midrule
$50$ & $50$ & $8$ & $28.7$ & $27.2$ & $15.1$ & $20.6$ & $19.2$ & $11.6$ \\
$100$ & $50$ & $8$ & $23.0$ & $22.3$ & $12.1$ & $20.6$ & $19.2$ & $11.6$ \\
$50$ & $100$ & $8$ & $28.3$ & $26.6$ & $14.7$ & $23.8$ & $21.1$ & $12.9$ \\
$100$ & $100$ & $8$ & $28.1$ & $26.1$ & $14.3$ & $25.1$ & $21.9$ & $13.1$ \\
$50$ & $150$ & $8$ & $28.2$ & $26.6$ & $14.7$ & $23.9$ & $21.1$ & $12.9$ \\
$100$ & $150$ & $8$ & $28.0$ & $26.0$ & $14.3$ & $25.0$ & $21.7$ & $13.0$ \\
$50$ & $50$ & $10$ & $28.8$ & $27.3$ & $15.1$ & $\mathbf{26.1}$ & $\mathbf{22.6}$ & $\mathbf{13.4}$ \\
$100$ & $50$ & $10$ & $27.9$ & $26.3$ & $14.5$ & $24.3$ & $21.9$ & $13.2$ \\
$50$ & $100$ & $10$ & $28.8$ & $27.2$ & $15.1$ & $23.8$ & $21.0$ & $12.8$ \\
$100$ & $100$ & $10$ & $23.1$ & $22.5$ & $12.1$ & $24.4$ & $21.7$ & $13.2$ \\
$50$ & $150$ & $10$ & $23.0$ & $22.4$ & $12.0$ & $24.4$ & $22.0$ & $13.3$ \\
$100$ & $150$ & $10$ & $27.7$ & $26.2$ & $14.4$ & $23.9$ & $21.1$ & $12.9$ \\
$50$ & $50$ & $12$ & $28.9$ & $27.5$ & $15.2$ & $24.8$ & $21.7$ & $12.9$ \\
$100$ & $50$ & $12$ & $24.6$ & $23.4$ & $12.5$ & $24.4$ & $21.8$ & $13.2$ \\
$50$ & $100$ & $12$ & $27.8$ & $26.3$ & $14.3$ & $24.4$ & $21.1$ & $12.7$ \\
$100$ & $100$ & $12$ & $\mathbf{29.0}$ & $\mathbf{27.5}$ & $\mathbf{15.2}$ & $25.6$ & $21.9$ & $13.1$ \\
$50$ & $150$ & $12$ & $28.7$ & $27.1$ & $14.8$ & $24.4$ & $21.1$ & $12.7$ \\
$100$ & $150$ & $12$ & $28.8$ & $27.3$ & $15.0$ & $24.6$ & $21.8$ & $13.2$ \\
        \bottomrule
    \end{tabular}
    \end{adjustbox}
    \label{tab:embedding_creation_ablation}
\end{table}
%
%
In \Cref{tab:embedding_creation_ablation}, we explore various values for Walk Count ($\WalkCount$), Walk Length ($\WalkLength$), and Window Size ($\WindowSizeNodeToVec$).
For egocentric videos, we achieve best results for $\WalkCount$ as $100$, $\WalkLength$ as $100$, and $\WindowSizeNodeToVec$ as $12$.
Due to the high variability of scenes in egocentric videos, the walk count required to update the embeddings are high.
This also leads to having a high walk length and a high number of frames in a window.
Instead, for third-person videos, we require comparatively less walk count ($50$), walk length ($50$), and window size ($10$).
As the majority of the videos in third-person datasets are from the internet, they skip a large number of repetitive portions of the task~\cite{sid_procel}. Due to this, the number of walks and the length are less.
Furthermore, these datasets contain multiple non-relevant frames (explanation/animation) that should be circumvented.

In \Cref{tab:graph_walk_ablation}, we analyses the return parameter that controls the likelihood of immediately revisiting a node in the walk, and in-out parameter that allows the search to differentiate between inward and outward nodes~\cite{node2vec-kdd2016}.
For both views, we obtain the highest results for $\ReturnParameter$ and $\InputParameter$ as $1.0$\dt{and $1.0$, respectively}.

\begin{table}[t]
    \centering
    \caption{\textbf{Hyper-parameters for walks over \meth}. Here, we perform multiple walks to analyse the hyper-parameters for \embdmeth.
    \textbf{R}, and \textbf{F} represent recall, and F-score, respectively\vspace{-1em}}
    \begin{adjustbox}{width=0.475\textwidth}
    \begin{tabular}{@{}llccccccc@{}}\toprule
\multirow{2}{1.2cm}{Return Parameter} & \multirow{2}{1.2cm}{In-out Parameter} & \multicolumn{3}{c}{PC Assembly} & \multicolumn{3}{c}{Change Tire}\\
        \cmidrule(lr){3-5}\cmidrule(l){6-8}
& & \textbf{R} & \textbf{F} & \textbf{IoU} & \textbf{R} & \textbf{F} & \textbf{IoU} \\
        \midrule
        $0.1$ & $0.5$ & $28.7$ & $26.9$ & $14.8$ & $24.5$ & $21.8$ & $13.2$ \\
$0.1$ & $1.0$ & $28.4$ & $26.7$ & $14.7$ & $24.5$ & $21.8$ & $13.2$ \\
$0.5$ & $0.1$ & $24.2$ & $23.0$ & $12.3$ & $20.5$ & $18.9$ & $11.4$ \\
$0.5$ & $1.0$ & $28.5$ & $26.8$ & $14.7$ & $20.6$ & $19.0$ & $11.5$ \\
$1.0$ & $0.1$ & $24.6$ & $24.6$ & $14.5$ & $\mathbf{25.6}$ & $\mathbf{21.9}$ & $13.0$ \\
$1.0$ & $0.5$ & $28.8$ & $27.3$ & $15.1$ & $20.4$ & $18.9$ & $11.4$ \\
$1.0$ & $1.0$ & $\mathbf{29.0}$ & $\mathbf{27.5}$ & $\mathbf{15.2}$ & $\mathbf{25.6}$ & $\mathbf{21.9}$ & $\mathbf{13.1}$ \\
        \bottomrule
    \end{tabular}
    \end{adjustbox}
    \label{tab:graph_walk_ablation}
\end{table}



%% file: sections/procedure_learning_ablation.tex
\Cref{tab:K_ablation} contains results obtained upon varying $K$ (number of key-steps) for the \framework framework.
The results follow the trend in the previous work~\cite{sid_procel} and are highest for $K = 7$.
The results decrease significantly as the values of $K$ increase.
\begin{table}[t]
    \centering
    \setlength{\tabcolsep}{9pt}
    \caption{\textbf{Tuning $K$}. 
    Here, the results are obtained for various values of $K$.
    \textbf{R}, and \textbf{F} represent recall, and F-score, respectively\vspace{-1em}}
    \begin{tabular}{@{}lcccccc@{}}\toprule
\multirow{2}{*}{$K$} & \multicolumn{3}{c}{PC Assembly} & \multicolumn{3}{c}{Change Tire}\\
        \cmidrule(lr){2-4}\cmidrule(l){5-7}
& \textbf{R} & \textbf{F} & \textbf{IoU} & \textbf{R} & \textbf{F} & \textbf{IoU} \\
        \midrule
        $7$ & $\mathbf{29.0}$ & $\mathbf{27.5}$ & $\mathbf{15.2}$ & $\mathbf{25.6}$ & $\mathbf{21.9}$ & $\mathbf{13.1}$ \\
        $10$ & $19.5$ & $20.4$ & $10.4$ & $17.6$ & $16.9$ & $10.2$ \\
        $12$ & $18.8$ & $20.5$ & $10.2$ & $14.4$ & $14.2$ & $8.5$ \\
        $15$ & $19.7$ & $22.5$ & $10.7$ & $13.5$ & $13.6$ & $7.4$ \\
        \bottomrule
    \end{tabular}
    \label{tab:K_ablation}
\end{table}

In \Cref{sec:background_frames}, we discussed the presence of a high number of background frames in procedure learning datasets~\cite{CrossTask}. 
To address this issue, we employ a hand-object interaction detection method on first-person videos. 
Frames that exhibit hand-object interaction are categorized as foreground, while the rest are considered background. 
The effectiveness of background frame filtration is demonstrated in \Cref{tab:background_ablation}, showcasing improved results for categories involving open spaces. 
For instance, in the Greek Salad dataset~\cite{egtea_gaze_p}, the subjects work in an unrestricted kitchen environment, as opposed to PC Assembly~\cite{sid_procel}, where they work in a confined space with hands being visible for the majority of the time.
It should be noted that hand-object detection in third-person videos fails due to small hands and constant hand visibility.
Hence, we exclusively utilize hand-object interaction for background filtering in egocentric videos.
The enhanced results for third-person videos are solely attributed to \framework.
Results for other categories in EgoProceL are in the supplementary materials.


\begin{table}[t]
    \centering
    \caption{\textbf{Detecting background frames}. Results are obtained upon filtering frames that do not contain hand-object interaction. Results improve for categories with subjects working in an unrestricted space. \textbf{R}, and \textbf{F} represent recall, and F-score, respectively\vspace{-1em}}
    \begin{adjustbox}{width=0.475\textwidth}
    \begin{tabular}{@{}lcccccc@{}}\toprule
\multirow{2}{2cm}{Hand-Object Interaction} & \multicolumn{3}{c}{PC Assembly} & \multicolumn{3}{c}{Greek Salad}\\
        \cmidrule(lr){2-4}\cmidrule(l){5-7}
& \textbf{R} & \textbf{F} & \textbf{IoU} & \textbf{R} & \textbf{F} & \textbf{IoU} \\
        \midrule
        Not Checked & $\mathbf{29.5}$ & $\mathbf{27.6}$ & $14.4$ & $25.4$ & $22.3$ & $12.7$ \\
        Checked & $29.0$ & $27.5$ & $\mathbf{15.2}$ & $\mathbf{34.9}$ & $\mathbf{26.5}$ & $\mathbf{21.4}$ \\
        \bottomrule
    \end{tabular}
    \end{adjustbox}
    \label{tab:background_ablation}
\end{table}

%% file: sections/video_count_ablation.tex
\Cref{tab:increasing_videos} contains results obtained by increasing the number of videos used to create \meth.
The objective here is to assess the effectiveness of the \framework framework in relation to the number of videos utilized.
Here, we specifically select tasks with a number of videos that are powers of two.
We then generate $\frac{\NumVideos}{\text{Video Count}}$ graphs and concatenate the results to maintain consistency with the other experiments.
As shown in \Cref{tab:increasing_videos}, the highest F-score and IoU are achieved when using the maximum number of videos.
This supports our main claim that employing \meth to create a unified representation for all task videos enables capturing both the temporal relationships within individual videos and the semantic relationships across multiple videos. 
Furthermore, as the dataset size increases, \framework, in conjunction with \meth, consistently achieves high-performance results.


\begin{table}[t]
    \centering
    \setlength{\tabcolsep}{5pt}
    \caption{\textbf{Number of Videos}. Here, the results are obtained upon systematically increasing the number of videos for creating \meth. \textbf{R}, and \textbf{F} represent recall, and F-score, respectively\vspace{-1em}}
    \begin{tabular}{@{}lcccccc@{}}\toprule
\multirow{2}{1.7cm}{Video Count} & \multicolumn{3}{c}{Bacon and Eggs~\cite{egtea_gaze_p}} & \multicolumn{3}{c}{Tie-Tie~\cite{joint_dynamic_summary}}\\
        \cmidrule(lr){2-4}\cmidrule(l){5-7}
& \textbf{R} & \textbf{F} & \textbf{IoU} & \textbf{R} & \textbf{F} & \textbf{IoU} \\
        \midrule
        $4$ & $23.6$ & $20.0$ & $11.4$ & $20.9$ & $18.4$ & $11.2$ \\
        $8$ & $25.0$ & $22.1$ & $12.1$ & $21.3$ & $18.8$ & $11.2$ \\
        $16$ & $\mathbf{27.8}$ & $\mathbf{23.1}$ & $\mathbf{12.6}$ & $20.1$ & $17.6$ & $10.7$ \\
        $32$ & $-$ & $-$ & $-$ & $20.2$ & $18.0$ & $10.8$ \\
        $64$ & $-$ & $-$ & $-$ & $\mathbf{23.5}$ & $\mathbf{19.7}$ & $\mathbf{11.4}$ \\
        \bottomrule
    \end{tabular}
    \label{tab:increasing_videos}
\end{table}

\mypara{Remark on hyper-parameter selection}
%
Determining the optimal values poses a challenge due to the unsupervised nature of the problem.
To address this, we conduct experiments using both first- and third-person views and perform an extensive ablation study.
We present a set of hyper-parameters for each view in \Cref{tab:hyperparameters}.
In an effort to achieve generalizable hyper-parameter tuning, we perform the ablation study on a single task from each view.
It is worth noting that the EgoProceL dataset~\cite{sid_procel} contains videos from multiple sources, resulting in significant domain variation.
Nonetheless, the superior performance of \framework over existing approaches on three datasets demonstrates that the chosen hyper-parameters are applicable across different datasets.

\mypara{\mt{Assumptions and Limitations}}
\mt{Proposed GPL framework exhibits limitations that stem from certain assumptions. Firstly, UnityGraph relies on subjects using similar objects for identical key-steps, which might lead to inaccuracies when dissimilar objects are employed. Additionally, while our method filters background frames based on hand-object interactions, cases like the `check booklet' from MECCANO challenge this assumption. We're actively working to address these limitations in future iterations of our work to enhance UnityGraph's robustness and accuracy.}

%% file: sections/conclusion.tex
Procedure learning is an important direction toward creating systems capable of assisting humans.
Contrary to current approaches, we propose the graph-based procedure learning (\framework) framework.
\framework consists of \meth that creates a unified representation for multiple videos of the same task.
\meth allows us to model both, temporal and spatial information.
The results obtained and the ablation performed demonstrate the capability of a graph-based approach for procedure learning.

\mypara{Acknowledgments} The work was supported in part by the Department of Science and Technology, Government of India, under DST/ICPS/Data-Science project ID T-138. The authors thank Makarand Tapaswi and Charu Sharma for their Topics in Deep Learning course which motivated the paper's central idea.

%% file: sections/supp_results_video_count.tex
\begin{table}[!hpb]
    \centering
    \setlength{\tabcolsep}{5pt}
    \caption{\textbf{Number of Videos}. The results are obtained upon systematically increasing the number of videos for creating \meth. Here, we compare the performance of \meth over CnC~\cite{sid_procel}. Numbers in \textbf{bold} represent the highest number in the column and \underline{underlined} number represent the highest number in the row for that metric. \textbf{R}, and \textbf{F} represent recall, and F-score, respectively}
    \begin{tabular}{@{}lcccccc@{}}\toprule
\multirow{2}{1.7cm}{Number of Videos} & \multicolumn{3}{c}{\meth (\textit{ours})} & \multicolumn{3}{c}{CnC~\cite{sid_procel}}\\
        \cmidrule(lr){2-4}\cmidrule(l){5-7}
& \textbf{R} & \textbf{F} & \textbf{IoU} & \textbf{R} & \textbf{F} & \textbf{IoU} \\
        \midrule
        $4$ & $\underline{23.6}$ & $\underline{20.0}$ & $\underline{11.4}$ & $18.8$ & $16.3$ & $4.4$\\
        $8$ & $\underline{25.0}$ & $\underline{22.1}$ & $\underline{12.1}$ & $20.3$ & $17.9$ & $6.1$\\
        $16$ & $\mathbf{\underline{27.8}}$ & $\mathbf{\underline{23.1}}$ & $\mathbf{\underline{12.6}}$ & $21.2$ & $18.6$ & $9.3$ \\
        \bottomrule
    \end{tabular}
    \label{tab:supp_increasing_videos}
\end{table}

%% file: sections/supp_results_background.tex
\begin{table*}[t]
    \centering
    \setlength{\tabcolsep}{7pt}
    \caption{\textbf{Detecting the background frames.} Here, the results are obtained upon filtering the frames that do not contain hand-object interaction. For the categories that involve the subject moving in an unrestricted space and performing the tasks, the results improve. For example, CMU-MMAC and EPIC-Tents.
    This shows that detecting hand-object interaction is an efficient technique for identifying the background frames.
    \mt{Note that the results even without removing the background frames improve over state-of-the-art method}~\cite{sid_procel}  \mt{for most of the categories.
    This highlights the efficacy of the proposed GPL framework}
    }
    \begin{tabular}{@{}lcccccccccccc@{}}
    \toprule
    & \multicolumn{12}{c}{EgoProceL}\\
    \cmidrule{2-13}
    & \multicolumn{2}{c}{CMU-MMAC} & \multicolumn{2}{c}{EGTEA G.} & \multicolumn{2}{c}{MECCANO} & \multicolumn{2}{c}{EPIC-Tents} & \multicolumn{2}{c}{PC Assembly} & \multicolumn{2}{c}{PC Disassembly} \\
    \cmidrule(lr){2-3}\cmidrule(lr){4-5}\cmidrule(lr){6-7}\cmidrule(lr){8-9}\cmidrule(lr){10-11}\cmidrule(l){12-13}
    & F1 & IoU & F1 & IoU & F1 & IoU & F1 & IoU & F1 & IoU & F1 & IoU \\
    \midrule
    \mt{CnC}~\cite{sid_procel} & \mt{$22.7$} & \mt{$11.1$} & \mt{$21.7$} & \mt{$9.5$} & \mt{$18.1$} & \mt{$7.8$} & \mt{$17.2$} & \mt{$8.3$} & \mt{$25.1$} & \mt{$12.8$} & $\textbf{\mt{27.0}}$ & \mt{$14.8$} \\
    Not Checked & $30.2$ & $16.7$ & $23.6$ & $14.9$ & $20.6$ & $9.8$ & $18.3$ & $8.5$ & $\mathbf{27.6}$ & $14.4$ & $26.9$ & $15.0$\\
    Checked & $\textbf{31.7}$ & $\textbf{17.9}$ & $\textbf{27.1}$ & $\textbf{16.0}$ & $\textbf{20.7}$ & $\textbf{10.0}$ & $\textbf{19.8}$ & $\textbf{9.1}$ & $27.5$ & $\mathbf{15.2}$ & $26.7$ & $\textbf{15.2}$\\
    \bottomrule
    \end{tabular}
    \label{tab:supp_results_background}
\end{table*}

%% file: sections/supp_fig_tsne_vis.tex
\begin{figure*}[!tphb]
    \centering
    \includegraphics[width=\textwidth]{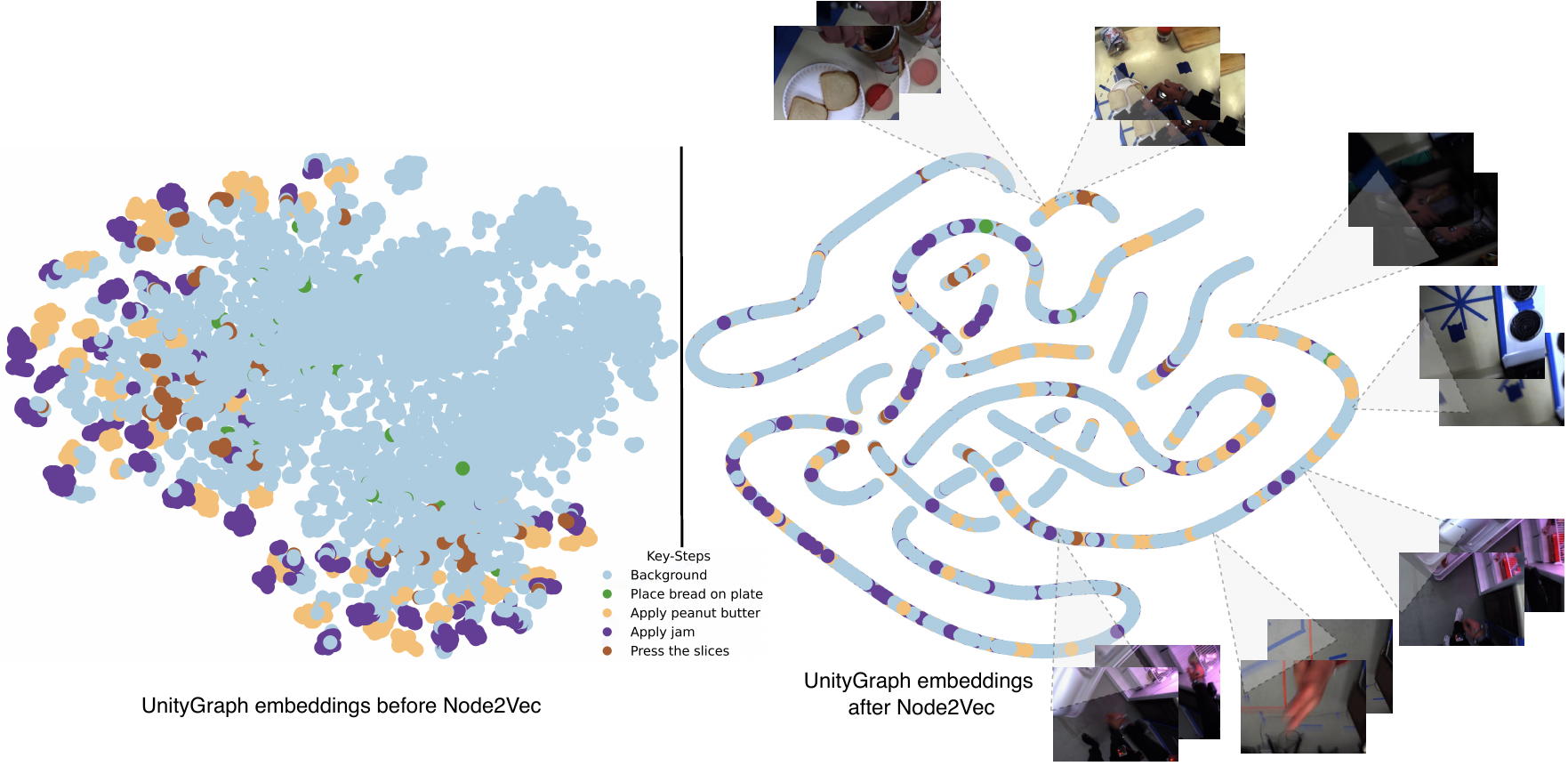}
    \caption{
    \textbf{t-SNE~\cite{tsne} visualisation} for the task of making a sandwich~\cite{CMU_Kitchens,sid_procel} before and after updating \meth's embeddings using the \embdmeth algorithm~\cite{node2vec-kdd2016}.
    Here, each color represents a key-step's category as noted in the legend.
    The left side of the figure consists of t-SNE visualisation obtained before using the \embdmeth algorithm.
    The right side of the figure consists of t-SNE visualisation obtained after updating \meth's embeddings using \embdmeth.
    As can be seen, upon updating the embeddings using Node2Vec, clips with similar key-steps come close.
    For example, the cluster on the top consists of clips of subjects applying peanut butter, whereas the cluster towards the centre has background clips of subject moving themselves from one place to other.
    }
    \label{fig:tsne_vis}
\end{figure*}

%% file: arxiv.bbl
\begin{thebibliography}{10}\itemsep=-1pt

\bibitem{Ahsan2018DiscrimNetSA}
U. Ahsan, Chen Sun, and Irfan Essa.
\newblock {DiscrimNet: Semi-Supervised Action Recognition from Videos using
  Generative Adversarial Networks}.
\newblock In {\em Computer Vision and Pattern Recognition Workshops (CVPRW)
  ‘Women in Computer Vision (WiCV)’}, 2018.

\bibitem{Inria_dataset}
Jean-Baptiste Alayrac, Piotr Bojanowski, Nishant Agrawal, Ivan Laptev, Josef
  Sivic, and Simon Lacoste-Julien.
\newblock {{Unsupervised learning from Narrated Instruction Videos}}.
\newblock In {\em Computer Vision and Pattern Recognition (CVPR)}, 2016.

\bibitem{sid_procel}
Siddhant Bansal, Chetan Arora, and C.V. Jawahar.
\newblock {My View is the Best View: Procedure Learning from Egocentric
  Videos}.
\newblock In {\em European Conference on Computer Vision (ECCV)}, 2022.

\bibitem{10.1007/978-3-319-10602-1_41}
Piotr Bojanowski, R{\'e}mi Lajugie, Francis Bach, Ivan Laptev, Jean Ponce,
  Cordelia Schmid, and Josef Sivic.
\newblock {Weakly Supervised Action Labeling in Videos under Ordering
  Constraints}.
\newblock In {\em European Conference on Computer Vision (ECCV)}, 2014.

\bibitem{Carreira2017QuoVA}
Jo{\~a}o Carreira and Andrew Zisserman.
\newblock {Quo Vadis, Action Recognition? A New Model and the Kinetics
  Dataset}.
\newblock In {\em Computer Vision and Pattern Recognition (CVPR)}, 2017.

\bibitem{Chang_2019_CVPR}
Chien-Yi Chang, De-An Huang, Yanan Sui, Li Fei-Fei, and Juan~Carlos Niebles.
\newblock {D3TW: Discriminative Differentiable Dynamic Time Warping for Weakly
  Supervised Action Alignment and Segmentation}.
\newblock In {\em Computer Vision and Pattern Recognition (CVPR)}, 2019.

\bibitem{chen2019graph}
Yunpeng Chen, Marcus Rohrbach, Zhicheng Yan, Yan Shuicheng, Jiashi Feng, and
  Yannis Kalantidis.
\newblock {Graph-based global reasoning networks}.
\newblock In {\em Proceedings of the IEEE/CVF Conference on Computer Vision and
  Pattern Recognition (CVPR)}, 2019.

\bibitem{idm}
Richard~W. Conners and Charles~A. Harlow.
\newblock {A Theoretical Comparison of Texture Algorithms}.
\newblock {\em IEEE Transactions on Pattern Analysis and Machine Intelligence},
  1980.

\bibitem{BMVC.28.30}
Dima Damen, Teesid Leelasawassuk, Osian Haines, Andrew Calway, and Walterio
  Mayol-Cuevas.
\newblock {You-Do, I-Learn: Discovering Task Relevant Objects and their Modes
  of Interaction from Multi-User Egocentric Video}.
\newblock In {\em British Machine Vision Conference (BMVC)}, 2014.

\bibitem{CMU_Kitchens}
F. De~La~Torre, J. Hodgins, A. Bargteil, X. Martin, J. Macey, A. Collado, and
  P. Beltran.
\newblock {Guide to the Carnegie Mellon University Multimodal Activity
  (CMU-MMAC) database.}
\newblock In {\em Robotics Institute}, 2008.

\bibitem{Diba2019DynamoNetDA}
Ali Diba, Vivek Sharma, L. Gool, and R. Stiefelhagen.
\newblock {DynamoNet: Dynamic Action and Motion Network}.
\newblock In {\em International Conference on Computer Vision (ICCV)}, 2019.

\bibitem{Ding2018WeaklySupervisedAS}
Li Ding and Chenliang Xu.
\newblock {Weakly-Supervised Action Segmentation with Iterative Soft Boundary
  Assignment}.
\newblock In {\em Computer Vision and Pattern Recognition (CVPR)}, 2018.

\bibitem{Doughty_2020_CVPR}
Hazel Doughty, Ivan Laptev, Walterio Mayol-Cuevas, and Dima Damen.
\newblock {Action Modifiers: Learning From Adverbs in Instructional Videos}.
\newblock In {\em Computer Vision and Pattern Recognition (CVPR)}, 2020.

\bibitem{tcc}
Debidatta Dwibedi, Yusuf Aytar, Jonathan Tompson, Pierre Sermanet, and Andrew
  Zisserman.
\newblock {Temporal Cycle-Consistency Learning}.
\newblock In {\em Computer Vision and Pattern Recognition (CVPR)}, 2019.

\bibitem{multi-task-procl}
Ehsan Elhamifar and Dat Huynh.
\newblock {{Self-supervised Multi-task Procedure Learning from Instructional
  Videos}}.
\newblock In {\em European Conference on Computer Vision (ECCV)}, 2020.

\bibitem{joint_dynamic_summary}
Ehsan Elhamifar and Zwe Naing.
\newblock {Unsupervised Procedure Learning via Joint Dynamic Summarization}.
\newblock In {\em International Conference on Computer Vision (ICCV)}, 2019.

\bibitem{Fernando2017SelfSupervisedVR}
Basura Fernando, Hakan Bilen, E. Gavves, and Stephen Gould.
\newblock {Self-Supervised Video Representation Learning with Odd-One-Out
  Networks}.
\newblock In {\em Computer Vision and Pattern Recognition (CVPR)}, 2017.

\bibitem{Fried2020LearningTS}
Daniel Fried, Jean-Baptiste Alayrac, P. Blunsom, Chris Dyer, S. Clark, and Aida
  Nematzadeh.
\newblock {Learning to Segment Actions from Observation and Narration}.
\newblock In {\em Association for Computational Linguistics (ACL)}, 2020.

\bibitem{node2vec-kdd2016}
Aditya Grover and Jure Leskovec.
\newblock {node2vec: Scalable Feature Learning for Networks}.
\newblock In {\em Proceedings of the 22nd ACM SIGKDD International Conference
  on Knowledge Discovery and Data Mining}, 2016.

\bibitem{networkx}
Aric~A. Hagberg, Daniel~A. Schult, and Pieter~J. Swart.
\newblock {Exploring Network Structure, Dynamics, and Function using NetworkX}.
\newblock In {\em Proceedings of the 7th Python in Science Conference}, 2008.

\bibitem{Han19dpc}
Tengda Han, Weidi Xie, and Andrew Zisserman.
\newblock {Video Representation Learning by Dense Predictive Coding}.
\newblock In {\em Workshop on Large Scale Holistic Video Understanding, ICCV},
  2019.

\bibitem{resnet}
Kaiming He, Xiangyu Zhang, Shaoqing Ren, and Jian Sun.
\newblock {Deep Residual Learning for Image Recognition}.
\newblock In {\em Computer Vision and Pattern Recognition (CVPR)}, 2016.

\bibitem{huang-buch-2018-finding-it}
De-An Huang*, Shyamal Buch*, Lucio Dery, Animesh Garg, Li Fei-Fei, and
  Juan~Carlos Niebles.
\newblock {Finding ``It'': Weakly-Supervised, Reference-Aware Visual Grounding
  in Instructional Videos}.
\newblock In {\em Proceedings of the IEEE/CVF Conference on Computer Vision and
  Pattern Recognition (CVPR)}, 2018.

\bibitem{Huang2016ConnectionistTM}
De-An Huang, Li Fei-Fei, and Juan~Carlos Niebles.
\newblock {Connectionist Temporal Modeling for Weakly Supervised Action
  Labeling}.
\newblock In {\em European Conference on Computer Vision (ECCV)}, 2016.

\bibitem{Huang2019NeuralTG}
De-An Huang, Suraj Nair, Danfei Xu, Yuke Zhu, Animesh Garg, Li Fei-Fei, Silvio
  Savarese, and Juan~Carlos Niebles.
\newblock {Neural Task Graphs: Generalizing to Unseen Tasks From a Single Video
  Demonstration}.
\newblock {\em Proceedings of the IEEE/CVF Conference on Computer Vision and
  Pattern Recognition (CVPR)}, 2019.

\bibitem{Hussein2019VideoGraphRM}
Noureldien Hussein, Efstratios Gavves, and Arnold W.~M. Smeulders.
\newblock {VideoGraph: Recognizing Minutes-Long Human Activities in Videos}.
\newblock {\em ArXiv}, abs/1905.05143, 2019.

\bibitem{9706996}
Lei Ji, Chenfei Wu, Daisy Zhou, Kun Yan, Edward Cui, Xilin Chen, and Nan Duan.
\newblock {Learning Temporal Video Procedure Segmentation from an Automatically
  Collected Large Dataset}.
\newblock In {\em 2022 IEEE/CVF Winter Conference on Applications of Computer
  Vision (WACV)}, 2022.

\bibitem{Khan2021SpatiotemporalDM}
Salman Khan and Fabio Cuzzolin.
\newblock {Spatiotemporal Deformable Models for Long-Term Complex Activity
  Detection}.
\newblock In {\em British Machine Vision Conference (BMVC)}, 2021.

\bibitem{Kim2019SelfSupervisedVR}
Dahun Kim, Donghyeon Cho, and In-So Kweon.
\newblock {Self-Supervised Video Representation Learning with Space-Time Cubic
  Puzzles}.
\newblock In {\em AAAI Conference on Artificial Intelligence}, 2019.

\bibitem{hungarian_algorithm}
H.~W. Kuhn.
\newblock {The Hungarian method for the assignment problem}.
\newblock {\em Naval Research Logistics Quarterly}, 1955.

\bibitem{kukleva2019unsupervised}
Anna Kukleva, Hilde Kuehne, Fadime Sener, and Jurgen Gall.
\newblock {Unsupervised learning of action classes with continuous temporal
  embedding}.
\newblock In {\em Computer Vision and Pattern Recognition (CVPR)}, 2019.

\bibitem{Kumar_2022_CVPR}
Sateesh Kumar, Sanjay Haresh, Awais Ahmed, Andrey Konin, M.~Zeeshan Zia, and
  Quoc-Huy Tran.
\newblock {Unsupervised Action Segmentation by Joint Representation Learning
  and Online Clustering}.
\newblock In {\em Proceedings of the IEEE/CVF Conference on Computer Vision and
  Pattern Recognition (CVPR)}, 2022.

\bibitem{Lee2017UnsupervisedRL}
Hsin-Ying Lee, Jia-Bin Huang, Maneesh~Kumar Singh, and Ming-Hsuan Yang.
\newblock {Unsupervised Representation Learning by Sorting Sequences}.
\newblock In {\em International Conference on Computer Vision (ICCV)}, 2017.

\bibitem{Li_2019_ICCV}
Jun Li, Peng Lei, and Sinisa Todorovic.
\newblock {Weakly Supervised Energy-Based Learning for Action Segmentation}.
\newblock In {\em International Conference on Computer Vision (ICCV)}, 2019.

\bibitem{Li_2020_CVPR}
Jun Li and Sinisa Todorovic.
\newblock {Set-Constrained Viterbi for Set-Supervised Action Segmentation}.
\newblock In {\em Computer Vision and Pattern Recognition (CVPR)}, 2020.

\bibitem{egtea_gaze_p}
Yin Li, Miao Liu, and James~M. Rehg.
\newblock {In the Eye of Beholder: Joint Learning of Gaze and Actions in First
  Person Video}.
\newblock In {\em European Conference on Computer Vision (ECCV)}, 2018.

\bibitem{8953941}
Xingyu Liu, Joon-Young Lee, and Hailin Jin.
\newblock {Learning Video Representations From Correspondence Proposals}.
\newblock In {\em Proceedings of the IEEE/CVF Conference on Computer Vision and
  Pattern Recognition (CVPR)}, 2019.

\bibitem{Malmaud2015WhatsCI}
J. Malmaud, Jonathan Huang, V. Rathod, Nick Johnston, Andrew Rabinovich, and K.
  Murphy.
\newblock {What's Cookin'? Interpreting Cooking Videos using Text, Speech and
  Vision}.
\newblock In {\em HLT-NAACL}, 2015.

\bibitem{Misra2016ShuffleAL}
Ishan Misra, C.~L. Zitnick, and M. Hebert.
\newblock {Shuffle and Learn: Unsupervised Learning Using Temporal Order
  Verification}.
\newblock In {\em European Conference on Computer Vision (ECCV)}, 2016.

\bibitem{procedure_completion_BMVC_2020}
Zwe Naing and Ehsan Elhamifar.
\newblock {Procedure Completion by Learning from Partial Summaries}.
\newblock In {\em British Machine Vision Conference (BMVC)}, 2020.

\bibitem{10.1007/978-3-031-19830-4_31}
Medhini Narasimhan, Arsha Nagrani, Chen Sun, Michael Rubinstein, Trevor
  Darrell, Anna Rohrbach, and Cordelia Schmid.
\newblock {TL;DW? Summarizing Instructional Videos with Task Relevance and
  Cross-Modal Saliency}.
\newblock In {\em European Conference on Computer Vision (ECCV)}, 2022.

\bibitem{PyTorch}
Adam Paszke, Sam Gross, Francisco Massa, Adam Lerer, James Bradbury, Gregory
  Chanan, Trevor Killeen, Zeming Lin, Natalia Gimelshein, Luca Antiga, Alban
  Desmaison, Andreas Kopf, Edward Yang, Zachary DeVito, Martin Raison, Alykhan
  Tejani, Sasank Chilamkurthy, Benoit Steiner, Lu Fang, Junjie Bai, and Soumith
  Chintala.
\newblock {PyTorch: An Imperative Style, High-Performance Deep Learning
  Library}.
\newblock In {\em Neural Information Processing Systems}, 2019.

\bibitem{deepwalk}
Bryan Perozzi, Rami Al-Rfou, and Steven Skiena.
\newblock {DeepWalk: Online Learning of Social Representations}.
\newblock In {\em Proceedings of the 20th ACM SIGKDD International Conference
  on Knowledge Discovery and Data Mining}, 2014.

\bibitem{Qian_2022_CVPR}
Yicheng Qian, Weixin Luo, Dongze Lian, Xu Tang, Peilin Zhao, and Shenghua Gao.
\newblock {SVIP: Sequence VerIfication for Procedures in Videos}.
\newblock In {\em Proceedings of the IEEE/CVF Conference on Computer Vision and
  Pattern Recognition (CVPR)}, 2022.

\bibitem{8578725}
Alexander Richard, Hilde Kuehne, and Juergen Gall.
\newblock {Action Sets: Weakly Supervised Action Segmentation Without Ordering
  Constraints}.
\newblock In {\em Computer Vision and Pattern Recognition (CVPR)}, 2018.

\bibitem{Richard_2018_CVPR}
Alexander Richard, Hilde Kuehne, Ahsan Iqbal, and Juergen Gall.
\newblock {NeuralNetwork-Viterbi: A Framework for Weakly Supervised Video
  Learning}.
\newblock In {\em Computer Vision and Pattern Recognition (CVPR)}, 2018.

\bibitem{Sener_2019_ICCV}
Fadime Sener and Angela Yao.
\newblock {Zero-Shot Anticipation for Instructional Activities}.
\newblock In {\em International Conference on Computer Vision (ICCV)}, 2019.

\bibitem{Sener_2015_ICCV}
Ozan Sener, Amir~R. Zamir, Silvio Savarese, and Ashutosh Saxena.
\newblock {Unsupervised Semantic Parsing of Video Collections}.
\newblock In {\em International Conference on Computer Vision (ICCV)}, 2015.

\bibitem{Shan20}
Dandan Shan, Jiaqi Geng, Michelle Shu, and David Fouhey.
\newblock Understanding human hands in contact at internet scale.
\newblock In {\em Proceedings of the IEEE/CVF Conference on Computer Vision and
  Pattern Recognition (CVPR)}, 2020.

\bibitem{Shen_action_segmentation_2021_CVPR}
Yuhan Shen, Lu Wang, and Ehsan Elhamifar.
\newblock {Learning To Segment Actions From Visual and Language Instructions
  via Differentiable Weak Sequence Alignment}.
\newblock In {\em Computer Vision and Pattern Recognition (CVPR)}, 2021.

\bibitem{10.5555/3045118.3045209}
Nitish Srivastava, Elman Mansimov, and Ruslan Salakhutdinov.
\newblock {Unsupervised Learning of Video Representations Using LSTMs}.
\newblock In {\em International Conference on Machine Learning (ICML)}, 2015.

\bibitem{tsne}
Laurens van~der Maaten and Geoffrey Hinton.
\newblock {Visualizing Data using t-SNE}.
\newblock {\em Journal of Machine Learning Research}, 2008.

\bibitem{VidalMata_2021_WACV}
Rosaura~G. VidalMata, Walter~J. Scheirer, Anna Kukleva, David Cox, and Hilde
  Kuehne.
\newblock {Joint Visual-Temporal Embedding for Unsupervised Learning of Actions
  in Untrimmed Sequences}.
\newblock In {\em Proceedings of the IEEE/CVF Winter Conference on Applications
  of Computer Vision (WACV)}, 2021.

\bibitem{10.5555/3157096.3157165}
Carl Vondrick, Hamed Pirsiavash, and Antonio Torralba.
\newblock {Generating Videos with Scene Dynamics}.
\newblock In {\em Neural Information Processing Systems}, 2016.

\bibitem{Wang_2021_WACV}
Shaojie Wang, Wentian Zhao, Ziyi Kou, Jing Shi, and Chenliang Xu.
\newblock {How to Make a BLT Sandwich? Learning VQA Towards Understanding Web
  Instructional Videos}.
\newblock In {\em Proceedings of the IEEE/CVF Winter Conference on Applications
  of Computer Vision (WACV)}, 2021.

\bibitem{10.1007/978-3-030-01228-1_25}
Xiaolong Wang and Abhinav Gupta.
\newblock Videos as space-time region graphs.
\newblock In {\em European Conference on Computer Vision (ECCV)}, 2018.

\bibitem{8578938}
Donglai Wei, Joseph Lim, Andrew Zisserman, and William~T Freeman.
\newblock {Learning and Using the Arrow of Time}.
\newblock In {\em Computer Vision and Pattern Recognition (CVPR)}, 2018.

\bibitem{Choi2020ShuffleAA}
Jin woo Choi, Gaurav Sharma, S. Schulter, and Jia-Bin Huang.
\newblock {Shuffle and Attend: Video Domain Adaptation}.
\newblock In {\em European Conference on Computer Vision (ECCV)}, 2020.

\bibitem{Xu_2019_CVPR}
Dejing Xu, Jun Xiao, Zhou Zhao, Jian Shao, Di Xie, and Yueting Zhuang.
\newblock {Self-Supervised Spatiotemporal Learning via Video Clip Order
  Prediction}.
\newblock In {\em Computer Vision and Pattern Recognition (CVPR)}, 2019.

\bibitem{xu2020gtad}
Mengmeng Xu, Chen Zhao, David~S. Rojas, Ali Thabet, and Bernard Ghanem.
\newblock {G-TAD: Sub-Graph Localization for Temporal Action Detection}.
\newblock In {\em Proceedings of the IEEE/CVF Conference on Computer Vision and
  Pattern Recognition (CVPR)}, 2020.

\bibitem{You_2022_WACV}
Chenyu You, Lianyi Han, Aosong Feng, Ruihan Zhao, Hui Tang, and Wei Fan.
\newblock {MEGAN: Memory Enhanced Graph Attention Network for Space-Time Video
  Super-Resolution}.
\newblock In {\em Proceedings of the IEEE/CVF Winter Conference on Applications
  of Computer Vision (WACV)}, 2022.

\bibitem{10.1145/2647868.2654997}
Shoou-I Yu, Lu Jiang, and Alexander Hauptmann.
\newblock {Instructional Videos for Unsupervised Harvesting and Learning of
  Action Examples}.
\newblock In {\em ACM International Conference on Multimedia}, 2014.

\bibitem{PGCN2019ICCV}
Runhao Zeng, Wenbing Huang, Mingkui Tan, Yu Rong, Peilin Zhao, Junzhou Huang,
  and Chuang Gan.
\newblock {Graph Convolutional Networks for Temporal Action Localization}.
\newblock In {\em International Conference on Computer Vision (ICCV)}, 2019.

\bibitem{Zhao_2022_CVPR}
He Zhao, Isma Hadji, Nikita Dvornik, Konstantinos~G. Derpanis, Richard~P.
  Wildes, and Allan~D. Jepson.
\newblock {P3IV: Probabilistic Procedure Planning From Instructional Videos
  With Weak Supervision}.
\newblock In {\em Proceedings of the IEEE/CVF Conference on Computer Vision and
  Pattern Recognition (CVPR)}, 2022.

\bibitem{YouCook2}
Luowei Zhou, Chenliang Xu, and Jason~J Corso.
\newblock {Towards Automatic Learning of Procedures From Web Instructional
  Videos}.
\newblock In {\em AAAI Conference on Artificial Intelligence}, 2018.

\bibitem{CrossTask}
Dimitri Zhukov, Jean-Baptiste Alayrac, Ramazan~Gokberk Cinbis, David Fouhey,
  Ivan Laptev, and Josef Sivic.
\newblock {Cross-task weakly supervised learning from instructional videos}.
\newblock In {\em Computer Vision and Pattern Recognition (CVPR)}, 2019.

\bibitem{zhukov20}
D. Zhukov, J.-B. Alayrac, I. Laptev, and J. Sivic.
\newblock {Learning Actionness via Long-range Temporal Order Verification}.
\newblock In {\em European Conference on Computer Vision (ECCV)}, 2020.

\end{thebibliography}
